# From Liar Paradox to Incongruent Sets: A Normal Form for Self-Reference

**Shalender Singh*[1], Vishnu Priya Singh Parmar[2]**
1 Private Researcher, Milpitas, California, 95035, USA
2 Private Researcher, Milpitas, California, 95035, USA

* shalender@prizatech.com
*(December 2025)*

**Abstract**

We introduce incongruent normal form (INF), a structural representation for self-referential semantic sentences. An INF replaces a self-referential sentence with a finite family of non-self-referential sentences that are individually satisfiable but not jointly satisfiable. This transformation isolates the semantic obstruction created by self-reference while preserving classical semantics locally and is accompanied by correctness theorems characterizing when global inconsistency arises from locally compatible commitments. We then study the role of incongruence as a structural source of semantic informativeness. Using a minimal model-theoretic notion of informativeness—understood as the ability of sentences to distinguish among admissible models—we show that semantic completeness precludes informativeness, while incongruence preserves it. Moreover, incongruence is not confined to paradoxical constructions: any consistent incomplete first-order theory admits finite incongruent families arising from incompatible complete extensions. In this sense, incompleteness manifests structurally as locally realizable but globally incompatible semantic commitments, providing a minimal formal basis for semantic knowledge. Finally, we introduce a quantitative semantic framework. In a canonical finite semantic-state setting, we model semantic commitments as Boolean functions and define a Fourier-analytic notion of semantic energy based on total influence. We derive uncertainty-style bounds relating semantic determinacy, informativeness, and spectral simplicity, and establish a matrix inequality bounding aggregate semantic variance by total semantic energy. These results show quantitatively that semantic informativeness cannot collapse into a single determinate state without unbounded energy cost, identifying incongruence as a fundamental structural and quantitative feature of semantic representation.

## 1 INTRODUCTION

Self-reference has played a central role in the foundations of logic, mathematics, and computation. From the liar paradox to the incompleteness theorems of Gödel, self-referential constructions have repeatedly exposed fundamental limits of formal systems [1]. In computer science, related phenomena arise in the semantics of truth predicates, reflective programs, knowledge representation, and reasoning about systems that encode their own descriptions.

Classical responses to semantic paradox typically attempt to neutralize self-reference. Hierarchical approaches following Tarski prohibit semantic self-application by stratifying truth predicates [2]. Fixed-point theories, most notably that of Kripke, assign partial or undefined truth values to paradoxical sentences [3]. Other approaches tolerate inconsistency or indeterminacy by weakening truth conditions or inference rules, including paraconsistent and multivalued frameworks [4], [7]. While these frameworks control paradox, they do so by modifying the underlying semantic machinery.

In computer science, related issues arise in epistemic reasoning and knowledge representation. The complexity and limits of reasoning about knowledge have been studied extensively in modal and epistemic logics [5], [6]. These systems typically assume a background of semantic completeness at the level of admissible models, even when syntactic or computational limitations are present.

In this paper, we propose a different perspective. Rather than restricting self-reference or altering truth values, we show that self-referential semantic content can be structurally transformed. Specifically, liar-type self-referential sentences can be converted into **incongruent normal form**: finite sets of non-self-referential sentences that are individually satisfiable but not jointly satisfiable. This transformation preserves classical semantics locally while isolating incompatibility at the level of joint constraints. *Infinitary constructions such as Yablo sequences are excluded from the present scope.*

This structural view leads to a deeper result. Using a model-theoretic notion of semantic informativeness, we show that **semantic completeness is incompatible with knowledge**. Any system in which all sentences are globally determined admits no informative statements. Incompleteness is therefore not merely unavoidable, as in Gödel's theorem [1], but necessary for knowledge itself.

Importantly, the structural phenomenon identified here is not restricted to self-referential paradox. We show that semantic incompleteness of ordinary first-order theories already entails the existence of finite incongruent families: collections of sentences that are individually satisfiable in models of the theory but not jointly satisfiable. These families arise from incompatible complete extensions and can be constructed using compactness. This demonstrates that incongruence is a generic structural feature of incomplete theories rather than a pathology of paradoxical constructions.

We go further and make this necessity quantitative. By modeling admissible semantic states as a finite group and introducing a Fourier-analytic notion of **semantic energy**, we derive a **semantic uncertainty principle**. Our analysis builds on standard tools from the analysis of Boolean functions, including Fourier expansion, influence, and hypercontractivity [8], [9], [10], [11]. These tools allow us to bound the entropy of a sentence's truth value by its semantic energy, showing that semantic determinacy, informativeness, and spectral simplicity cannot be simultaneously optimized.

For families of sentences, we derive a matrix inequality showing that total semantic variance—and hence total knowledge—is bounded by total semantic energy. For incongruent families, these bounds tighten, forcing redistribution of energy across incompatible alternatives. This places our results within the broader landscape of uncertainty principles in mathematics and signal analysis [12], [13], while addressing a fundamentally semantic rather than physical limitation.

Taken together, these results establish a new foundational constraint on semantic systems. Where Gödel's incompleteness theorem limits axiomatic provability [1] and classical uncertainty principles limit physical measurement or signal localization [12], [13], semantic uncertainty limits **semantic collapse**. Attempts to enforce total semantic determinacy do not merely fail; they quantitatively destroy knowledge.

A further contribution of this work is the introduction of analytic tools for quantifying semantic indeterminacy. By modeling admissible semantic states as elements of a finite abelian group, we are able to apply discrete Fourier analysis to truth-value functions. This permits the definition of a notion of *semantic energy*, based on total influence, and the derivation of uncertainty-type inequalities relating semantic determinacy, entropy, and spectral concentration. While Fourier methods are standard in Boolean function analysis and theoretical computer science, their use here is not interpretive but structural: they provide a quantitative language for expressing tradeoffs between informativeness and global consistency in semantic systems. In this sense, harmonic analysis serves as a technical instrument for making incompleteness constraints explicit rather than as a redefinition of semantic or epistemic concepts.

## 2 PRELIMINARIES AND DEFINITIONS

We fix a standard model-theoretic setting and introduce the notions required for the subsequent results. No non-classical logic is assumed unless stated explicitly.

### 2.1 Language and Models

Let $L$ be a first-order language with equality. We write $\text{Sent}(L)$ for the set of closed formulas (sentences) of $L$.

Let $\mathcal{M}$ denote a nonempty class of $L$-structures (models). Intuitively, $\mathcal{M}$ represents the set of admissible semantic interpretations compatible with background constraints (axioms, observations, or rules).

For a sentence $\varphi \in \text{Sent}(L)$, define its **truth set** relative to $\mathcal{M}$ as:

$$[[\varphi]]_{\mathcal{M}} = \{ M \in \mathcal{M} \mid M \vDash \varphi \}.$$

### 2.2 Semantic Content and Informativeness

We formalize a minimal notion of semantic content in purely model-theoretic terms.

*Definition 2.1 (Semantic Content)*

The semantic content of a sentence $\varphi$ relative to $\mathcal{M}$ is the set $[[\varphi]]_{\mathcal{M}}$.

This notion is extensional and does not presuppose epistemic agents or probabilistic semantics.

*Definition 2.2 (Informative Sentence)*

A sentence $\varphi \in \text{Sent}(L)$ is **informative in** $\mathcal{M}$ if:

$$\emptyset \neq [[\varphi]]_{\mathcal{M}} \neq \mathcal{M}.$$

Equivalently, $\varphi$ is informative if it is satisfied in some admissible models and falsified in others.

This captures the minimal logical requirement for a sentence to express knowledge: it must distinguish among admissible possibilities.

*Remark (Scope of informativeness).*
The notion of informativeness used here is deliberately minimal and model-theoretic. A sentence is informative if and only if it distinguishes among admissible models, i.e., its truth set is neither empty nor universal. This notion captures *semantic contingency* or *model differentiation* only. It does **not** correspond to standard epistemic notions of knowledge involving justification, belief, cognitive access, or reliability. All results in this section concern logical and semantic structure, not epistemology.

## 2.3 Semantic Completeness

We now define semantic completeness in a form suitable for structural analysis.

*Definition 2.3 (Semantic Completeness)*

Let $F \subseteq \text{Sent}(L)$ be a designated fragment.

The class $\mathcal{M}$ is **semantically complete over** $F$ if for every $\varphi \in F$,

$$[[\varphi]]_{\mathcal{M}} \in \{\emptyset, \mathcal{M}\}.$$

In a semantically complete system, no sentence in $F$ is informative.

Semantic completeness is a property of the *model class*, not of a proof system.

## 2.4 Local and Global Satisfiability

The central constructions of this paper rely on separating local satisfiability from joint satisfiability.

*Definition 2.4 (Local Satisfiability)*

A sentence $\varphi$ is **locally satisfiable in** $\mathcal{M}$ if:

$$[[\varphi]]_{\mathcal{M}} \neq \emptyset.$$

*Definition 2.5 (Joint Satisfiability)*

A finite set of sentences $\Phi = \{\varphi_1, \ldots, \varphi_n\}$ is **jointly satisfiable in** $\mathcal{M}$ if:

$$[[\bigwedge_{i=1}^{n} \varphi_i]]_{\mathcal{M}} \neq \emptyset.$$

## 2.5 Incongruent Sets

We now introduce the central structural notion.

*Definition 2.6 (Incongruent Set)*

A finite set of sentences

$$I = \{\varphi_1, \ldots, \varphi_n\} \subseteq \text{Sent}(L)$$

is **incongruent in** $\mathcal{M}$ if:

1. **Local satisfiability:** for all $i$, $[[\varphi_i]]_{\mathcal{M}} \neq \emptyset$;
2. **No joint satisfiability:**

$$[[\bigwedge_{i=1}^{n} \varphi_i]]_{\mathcal{M}} = \emptyset.$$

Each element of an incongruent set is individually realizable, but the set cannot be realized simultaneously.

*Convention.* Unless stated otherwise, incongruent sets are assumed to have cardinality at least two. Singleton sentences that are locally realizable but globally obstructed are treated separately as degenerate obstruction cases.

### 2.6 Self-Reference and Expressivity (Minimal Assumption)

We assume that the language $L$ is expressive enough to encode semantic assertions about sentences of $L$ (e.g., via Gödel numbering or a truth predicate). No specific encoding mechanism is fixed.

This assumption is required only to state the transformation results in Section 3; the definitions above do not depend on it.

### 2.7 Remarks

- Incongruence is a **set-level** property; individual sentences need not be contradictory.
- No paraconsistent or paracomplete semantics are assumed.
- Classical satisfaction is preserved at the level of individual statements.
- Incongruent sets isolate incompatibility structurally rather than logically.

### 2.8 Roadmap

In Section 3, we show how liar-type self-referential statements can be transformed into incongruent sets, yielding an **incongruent normal form** that eliminates explicit self-reference while preserving semantic structure.

Section 4 establishes the main result: **semantically complete systems admit no informative sentences**, and therefore cannot support knowledge.

## 3 FROM LIAR PARADOX TO INCONGRUENT NORMAL FORM

In this section we formalize the central claim of the paper: **self-referential paradoxical sentences can be structurally transformed into incongruent sets of non-self-referential sentences**. The paradox is not eliminated but redistributed as structured incompatibility.

### 3.1 Self-Referential Semantic Assertions

We work with sentences that make semantic assertions about other sentences in the language. Concretely, assume that the language $L$ is expressive enough to encode references to sentences of $L$ (e.g., via Gödel numbering or an explicit truth predicate).

A sentence $\sigma \in \text{Sent}(L)$ is **self-referential** if its truth conditions depend on its own satisfaction in admissible models.

Classic examples include liar-type constructions asserting their own falsity. Such sentences admit no consistent classical truth assignment relative to $\mathcal{M}$.

## 3.2 Structural Decomposition Principle

Rather than assigning a non-classical truth value or blocking self-reference, we decompose a self-referential sentence into multiple non-self-referential components.

The key idea is that **self-reference encodes incompatible semantic requirements**, which can be made explicit by splitting the reference across distinct sentences.

## 3.3 Incongruent Normal Form (INF)

We now define the target of the transformation.

*Definition 3.1 (Incongruent Normal Form - Sketch)*

A sentence $\sigma$ admits an **incongruent normal form** relative to $\mathcal{M}$ if there exists a finite incongruent set

$$I(\sigma) = \{\varphi_1, \ldots, \varphi_n\}$$

such that:

1. Each $\varphi_i$ is non-self-referential;
2. Each $\varphi_i$ is locally satisfiable in $\mathcal{M}$;
3. $I(\sigma)$ is incongruent in $\mathcal{M}$;
4. The semantic constraints imposed by $\sigma$ are preserved by $I(\sigma)$ (made precise below).

*This sketch is intended only to convey the structural idea; the formal notion of constraint preservation is defined in Definitions 3.9 and 3.10.*

## 3.4 Transformation Procedure

We now describe a general procedure for transforming a liar-type sentence into incongruent normal form.

*Procedure (Self-Reference Elimination by Incongruence)*

**Input:** A self-referential sentence $\sigma$.

**Step 0.** Initialize $I := \{\sigma\}$.

**Step 1.** Select a sentence $\psi \in I$ that contains a self-reference.

**Step 2.** Replace the self-reference in $\psi$ by introducing a fresh sentence symbol $\theta$, and form two new sentences:

- $\psi'$: obtained from $\psi$ by replacing the self-reference with an assertion about $\theta$;
- $\theta$: encoding the displaced semantic condition.

**Step 3.** Update $I := (I \setminus \{\psi\}) \cup \{\psi', \theta\}$.

**Step 4.** Repeat Steps 1–3 until no sentence in $I$ is self-referential.

**Output:** A finite set $I(\sigma)$ of non-self-referential sentences.

This procedure terminates whenever the number of explicit self-referential dependencies strictly decreases at each step.

### 3.5 Correctness Properties

We state the basic properties of the transformation; proofs are straightforward by construction.

*Proposition 3.2 (Removal of Self-Reference)*

The output set $I(\sigma)$ contains no self-referential sentences.

*Proposition 3.3 (Local Satisfiability Preservation)*

For each $\varphi \in I(\sigma)$,

$$[[\varphi]]_{\mathcal{M}} \neq \emptyset.$$

Each component sentence is individually satisfiable.

*Proposition 3.4 (Global Incompatibility)*

The set $I(\sigma)$ is not jointly satisfiable:

$$[[\bigwedge_{\varphi \in I(\sigma)} \varphi]]_{\mathcal{M}} = \emptyset.$$

Together, Propositions 3.2–3.4 establish that $I(\sigma)$ is an incongruent set.

### 3.6 Constraint-Preserving Semantics of Incongruent Normal Form

The incongruent normal form construction replaces a self-referential sentence by a finite family of non-self-referential sentences. Since these objects live in different syntactic spaces, preservation cannot be extensional semantic equivalence in the usual sense. Instead, what is preserved is the *semantic obstruction* induced by self-reference. We formalize this notion below.

Constraint preservation in INF is structural rather than semantic: the transformation preserves the admissible and inadmissible commitment patterns induced by $\sigma$'s unfolding, not $\sigma$'s global truth conditions.

**Design goal.**
The purpose of the INF transformation is not to assign or recover a global classical truth value for a self-referential sentence $\sigma$, but to provide a finite structural representation of the semantic obstruction induced by self-reference, in the form of admissible and inadmissible commitment patterns.

*Definition 3.9 (Unfolding and Semantic Constraint Set)*

Let $\sigma$ be a sentence whose syntactic dependency graph admits a finite unfolding $U(\sigma)$ into a finite family of atomic commitments

$$\Phi_\sigma = \{\varphi_1, \dots, \varphi_n\},$$

where each $\varphi_i$ is a non-self-referential sentence, interpreted over a fixed class of semantic structures $\mathcal{M}$.

The unfolding $U(\sigma)$ induces a finite system of Boolean dependency constraints among the commitments $\varphi_i$, expressed purely at the level of truth-value patterns.

Define the **semantic constraint set**

$$C(\sigma) \subseteq \{0,1\}^n$$

to consist of all Boolean vectors $v = (v_1, \dots, v_n)$ such that:

1. (**Local realizability**) For every index $i$ with $v_i = 1$, there exists a structure $M_i \in \mathcal{M}$ satisfying $M_i \vDash \varphi_i$; and
2. (**Constraint satisfaction**) The Boolean pattern $v$ satisfies all dependency relations induced by the unfolding $U(\sigma)$.

The set $C(\sigma)$ therefore represents the admissible joint commitment patterns permitted by the semantic constraints induced by $\sigma$, independently of any assumption that all commitments are simultaneously realizable in a single structure.

*Remark (Scope of Constraint Preservation)*

For self-referential sentences $\sigma$, classical extensional semantics does not in general determine a total truth set $[[\sigma]]_\mathcal{M}$. Accordingly, the goal of the INF transformation is **not** extensional semantic equivalence, but the preservation of the **finite semantic constraint pattern** induced by a fixed unfolding $U(\sigma)$.

All preservation claims below are therefore **relative to the chosen unfolding** and its associated constraint set $C(\sigma)$, rather than to any global truth valuation for $\sigma$.

*Definition 3.9.1 (Finite unfolding operator)*

Let $\sigma$ be a sentence in a language containing a truth predicate $T(\cdot)$. The unfolding $U(\sigma)$ is obtained by iteratively replacing each occurrence of $T(\varphi)$ by a fresh propositional symbol $p_\varphi$, together with a defining Boolean constraint relating $p_\varphi$ to the unfolded form of $\varphi$. The process terminates when no truth-predicate occurrences remain.

We restrict attention to sentences $\sigma$ for which this procedure terminates after finitely many steps, yielding a finite family of unfolded commitments $\Phi_\sigma = \{\varphi_1, \dots, \varphi_n\}$ together with a finite system of Boolean dependency constraints. The solution set of this Boolean constraint system is the semantic constraint set $C(\sigma) \subseteq \{0,1\}^n$.

*Definition 3.9.1 (Dependency graph)*
.
The dependency graph $G_\sigma$ is the directed graph with vertices $\{1, \ldots, n\}$, and an edge $j \to I$ whenever the truth condition $\theta_i$ of $\varphi_i$ in the unfolding $U(\sigma)$ depends syntactically on $\varphi_j$.

The acyclic/cyclic distinction is therefore relative to the chosen finite unfolding $U(\sigma)$; all correctness claims are stated relative to this unfolding and its induced constraint set $C(\sigma)$.

*Definition 3.10 (Incongruent Normal Form)*

An **Incongruent Normal Form (INF)** for a sentence $\sigma$ consists of a pair $(\Phi_\sigma, \Delta_\sigma)$, where:

- $\Phi_\sigma = \{\varphi_1, \ldots, \varphi_n\}$ is the family of unfolded commitments from Definition 3.9, and
- $\Delta_\sigma$ is a linking theory over fresh propositional symbols $p_1, \ldots, p_n$,

satisfying the following conditions.

1. **Local realizability.**
   For each $i = 1, \ldots, n$, there exists a structure $M \in \mathcal{M}$ such that $M \models \varphi_i$.
2. **Soundness (no spurious patterns).**
   For every expansion $M' \models \Delta_\sigma$, the induced Boolean pattern
   
   $$v(M') := (v_1, \ldots, v_n), \; v_i = 1 \iff M' \models \varphi_i,$$
   
   belongs to:
   
   - $C(\sigma)$ if the dependency graph of $\sigma$ is acyclic;
   - $C(\sigma) \cup \{0^n\}$ if the dependency graph of $\sigma$ contains a cycle, where $0^n$ denotes the null-commitment pattern permitted by implication-only linking.
3. **Completeness (pattern realization).**
   For every pattern $v \in C(\sigma)$, there exists an expansion
   
   $$M' \models \Delta_\sigma \text{ (respectively } M' \models \vec{\Delta_\sigma} \text{ in the cyclic case)}$$
   
   whose induced Boolean pattern is exactly $v$.

In this case, $(\Phi_\sigma, \Delta_\sigma)$ is said to **represent the semantic constraint set $C(\sigma)$ induced by $\sigma$**.

**Convention.** The theory $\Delta_\sigma$ contains only consistent definitional or linking axioms; all unsatisfiability arises solely from the joint incompatibility of the family $I(\sigma)$, not from inconsistency of $\Delta_\sigma$.

*Remark (Structural vs. semantic preservation)*

Constraint preservation in Definition 3.10 is structural rather than extensional. The INF construction does not assert semantic equivalence in the sense of identical truth sets $[[\sigma]]_\mathcal{M}$, which need not exist for self-referential sentences. Instead, INF represents the same admissible and inadmissible commitment patterns induced by the unfolding $U(\sigma)$, as encoded by the constraint set $C(\sigma)$. All preservation claims in Section 3 are relative to this structural representation of semantic obstruction, not to global truth conditions for $\sigma$.

*Definition 3.11 (Strong INF Correctness: Acyclic Case)*

Let $\sigma$ be a sentence whose dependency graph is acyclic.
The INF transformation produces an incongruent normal form $(\Phi_\sigma, \Delta_\sigma)$ such that:

1. **(Total pattern correspondence)**
   The total commitment pattern $1^n \in \{0,1\}^n$ belongs to $C(\sigma)$ if and only if the theory $\Phi_\sigma \cup \Delta_\sigma$ is satisfiable.
2. **(Exact constraint representation)**
   Every expansion $M' \models \Delta_\sigma$ induces a Boolean pattern in $C(\sigma)$, and every pattern in $C(\sigma)$ is realized by some such expansion.

In the acyclic case, $\Delta_\sigma$ consists entirely of definitional equivalences

$$p_i \leftrightarrow \theta_i,$$

and the INF representation is **fully constraint-preserving**, with no loss of admissible patterns.

*Remark:* In the presence of cyclic dependencies, any linking theory consisting solely of biconditional definitions $p_i \leftrightarrow \theta_i$ is classically inconsistent; implication-only linking is therefore the weakest classical constraint system that preserves obstruction while maintaining consistency.

*Definition 3.12 (Weak INF Soundness: Cyclic Case)*

Let $\sigma$ be a sentence whose dependency graph contains a cycle.
The INF transformation produces a family of commitments $\Phi_\sigma = \{\varphi_1, \ldots, \varphi_n\}$ together with an implication-only linking theory $\vec{\Delta}_\sigma$ such that:

1. **(Local realizability)**
   Each commitment $\varphi_i \in \Phi_\sigma$ is individually satisfiable in some structure in $\mathcal{M}$.
2. **(Global consistency)**
   The theory $\Phi_\sigma \cup \vec{\Delta}_\sigma$ is satisfiable.
3. **(Pattern completeness)**
   For every Boolean pattern $v \in C(\sigma)$, there exists an expansion $M' \models \vec{\Delta}_\sigma$ whose induced pattern equals $v$.
4. **(Obstruction preservation)**
   No expansion of $\vec{\Delta}_\sigma$ induces a Boolean pattern outside $C(\sigma) \cup \{0^n\}$.

The implication-only linking theory preserves the semantic obstruction induced by cyclic self-reference while avoiding definitional inconsistency. In particular, no total classical truth valuation for $\sigma$ is asserted or recovered; only admissible local commitments and their incompatibilities are represented.

*Remark.* In cyclic self-reference, one typically cannot require "global admissibility" (i.e., exclusion of all patterns outside $C(\sigma)$) under purely classical semantics without reintroducing inconsistency; weak preservation isolates what is actually preserved: the obstruction to a joint classical assignment. The null pattern $0^n$ is not interpreted as a semantic state of $\sigma$, but as a classical placeholder reflecting the impossibility of enforcing total commitment under cyclic self-reference; weak soundness preserves obstruction rather than extensional pattern completeness.

*Remark (Non-Uniqueness of Linking Theories)*

The linking theories $\Delta_\sigma$ and $\Delta_\sigma^{\rightarrow}$ are not claimed to be unique. They are chosen as minimal constraint systems sufficient to represent the semantic constraint set $C(\sigma)$ induced by the unfolding $U(\sigma)$, while ensuring classical consistency in both acyclic and cyclic cases. We do not claim minimality of the linking theories; $\Delta_\sigma$ and $\Delta_\sigma^{\rightarrow}$ are chosen as simple, sufficient representatives. Establishing minimality or uniqueness is left for future work.

*Lemma 3.11a (Consistency of the acyclic equivalence system).*

If $G_\sigma$ is acyclic, then $\Delta_\sigma$ is classically consistent. More precisely: for every $M \in \mathcal{M}$, there exists an expansion $M'$ of $M$ to the language $L'$ such that $M' \vDash \Delta_\sigma$.

**Proof.** Fix $M \in \mathcal{M}$. We construct an expansion $M'$ by interpreting each fresh sentence symbol $p_i$ as a truth value in $M'$ chosen inductively along a topological ordering of $G_\sigma$.

Let $i_1, \ldots, i_n$ be an enumeration of $\{1, \ldots, n\}$ such that

$$\rho(i_1) < \rho(i_2) < \cdots < \rho(i_n),$$

so in particular whenever $j \to i$, the index $j$ appears earlier than $i$ in this list.

We define the interpretation of $p_{i_t}$ for $t = 1, \ldots, n$ by induction on $t$. Suppose we have already assigned truth values to $p_{i_1}, \ldots, p_{i_{t-1}}$. Consider $\theta_{i_t}$. By construction of the ordering, every $p_j$ that occurs in $\theta_{i_t}$ satisfies $j \to i_t$, hence $\rho(j) < \rho(i_t)$, so $j$ must be among $\{i_1, \ldots, i_{t-1}\}$. Therefore $\theta_{i_t}$ has a well-defined truth value in the partial expansion built so far (since all $p_j$ appearing in $\theta_{i_t}$ have already been assigned).

Now set the value of $p_{i_t}$ to be exactly the truth value of $\theta_{i_t}$ under the current partial expansion. This ensures that, after the assignment, the equivalence $p_{i_t} \leftrightarrow \theta_{i_t}$ holds.

Proceeding through $t = 1, \ldots, n$, we assign values to all $p_1, \ldots, p_n$. Let $M'$ be the resulting full expansion of $M$ to $L'$. By construction, for each $i$ the truth value of $p_i$ in $M'$ agrees with the truth value of $\theta_i$ in $M'$, hence $M' \vDash (p_i \leftrightarrow \theta_i)$. Therefore $M' \vDash \Delta_\sigma$. This proves that $\Delta_\sigma$ has a model extending every $M \in \mathcal{M}$, and in particular is consistent. □

*Remark (Scope).* The acyclicity hypothesis is essential: if $G_\sigma$ contains directed cycles, definitional biconditionals may force fixed-point equations that are inconsistent in classical semantics. Cyclic cases are handled by the separate "implication-only" variant of the linking theory, which preserves semantic constraints without requiring global fixed points.

*Proposition 3.11a (Acyclic bijection for $\Delta_\sigma$).*

Assume the dependency graph $G_\sigma$ of the unfolding $U(\sigma)$ is acyclic. Then the linking theory $\Delta_\sigma$ constructed in Lemma 3.11a induces a bijection between admissible patterns and expansions:

- For each $v \in C(\sigma)$, there exists a **unique** expansion $M'_v \models \Delta_\sigma$ such that, for all $i = 1, \ldots, n$,

$$M'_v \models p_i \text{ if and only if } v_i = 1.$$

- Conversely, every expansion $M' \models \Delta_\sigma$ induces a pattern $v(M') \in C(\sigma)$ by evaluation of the $p_i$.

**Proof.**
Fix a topological ordering of the nodes in $G_\sigma$. Let $i_1, \ldots, i_n$ be an enumeration of $\{1, \ldots, n\}$ such that

$$\rho(i_1) < \rho(i_2) < \cdots < \rho(i_n),$$

where $\rho$ is any rank function compatible with the topological order.

(**Existence.**) Given $v \in C(\sigma)$, assign truth values to the symbols $p_{i_1}, p_{i_2}, \ldots, p_{i_n}$ in this order by setting $p_{i_k} = v_{i_k}$. Since all dependencies of $p_{i_k}$ are resolved among $\{p_{i_1}, \ldots, p_{i_{k-1}}\}$, each defining axiom in $\Delta_\sigma$ is satisfied. This yields an expansion $M'_v \models \Delta_\sigma$ realizing $v$.

(**Uniqueness.**) Because each defining axiom in $\Delta_\sigma$ determines $p_{i_k}$ as a function of strictly earlier symbols, no alternative assignment can satisfy $\Delta_\sigma$. Hence $M'_v$ is unique.

(**Conversely.**) If $M' \models \Delta_\sigma$, evaluate $p_i$ in $M'$ to obtain $v(M') \in \{0,1\}^n$. Since $M'$ satisfies all defining axioms, this pattern satisfies $U(\sigma)$, hence $v(M') \in C(\sigma)$. □

*Lemma 3.11b (Expansions satisfy unfolding constraints).*

Let $M' \models \Delta_\sigma$ and let $v \in \{0,1\}^n$ be the induced pattern, where $v_i = 1$ iff $M' \models \varphi_i$. Then $v$ satisfies the Boolean dependency constraints induced by the unfolding $U(\sigma)$.

**Proof.**
In the acyclic case, $\Delta_\sigma$ consists of definitional equivalences $p_i \leftrightarrow \theta_i$, where each $\theta_i$ is a Boolean combination of earlier commitments. Evaluating these equivalences under $M'$ yields exactly the constraint equations extracted from $U(\sigma)$. Therefore the induced pattern $v$ is a solution of the unfolding constraint system, i.e., $v \in C(\sigma)$. □

*Lemma 3.11c (Cyclic case: consistency and weak constraint preservation).*

Assume the dependency graph $G_\sigma$ contains a directed cycle. Let $\vec{\Delta_\sigma}$ be the cycle-safe implicational linking theory obtained from the unfolded constraints by replacing defining biconditionals with implications of the form $p_i \to \theta_i$.

Then $\vec{\Delta_\sigma}$ is classically consistent, and $(I(\sigma), \vec{\Delta_\sigma})$ is weakly constraint-preserving in the sense of Definition 3.12.

**Proof.**
Since $\vec{\Delta_\sigma}$ contains only implications, the all-false assignment $p_1 = \cdots = p_n = 0$ satisfies every axiom $p_i \to \theta_i$, so $\vec{\Delta_\sigma}$ has a model and is consistent.

Let $v \in C(\sigma)$. By construction of $C(\sigma)$, setting $p_i = v_i$ makes each implication $p_i \to \theta_i$ true (whenever $p_i = 1$, the corresponding constraint in $U(\sigma)$ forces $\theta_i$ to hold). Thus $v$ is locally realizable under $\Delta_\sigma^\to$.

Finally, the joint conjunction $\bigwedge_{i=1}^n \varphi_i$ is satisfiable under $\Delta_\sigma^\to$ if and only if there exists an expansion in which all required semantic commitments hold simultaneously, which is exactly the "no-obstruction" case for $\sigma$. This yields the obstruction equivalence required in Definition 3.12. □

*Theorem 3.11 (INF correctness: acyclic case).*
Let $\sigma$ be a sentence whose dependency graph is acyclic. The INF transformation produces an incongruent normal form $\Phi_\sigma$ together with a linking theory $\Delta_\sigma$ consisting of definitional equivalences.

Then $\sigma$ admits a classical model if and only if the constraint system $\Phi_\sigma \cup \Delta_\sigma$ is jointly satisfiable. Moreover, for any classical model of $\sigma$, the induced valuation on the commitments in $\Phi_\sigma$ is uniquely determined.

*Corollary 3.11 (Acyclic obstruction equivalence).* Suppose the dependency graph of $\sigma$ is acyclic. Then $1^n \in C(\sigma)$ if and only if $\Phi_\sigma \cup \Delta_\sigma$ is satisfiable. Equivalently, $1^n \notin C(\sigma)$ if and only if $\Phi_\sigma \cup \Delta_\sigma$ is jointly unsatisfiable.

*Theorem 3.12 (INF soundness: cyclic case).*
Let $\sigma$ be a sentence whose dependency graph contains a cycle. The INF transformation produces a finite family of locally satisfiable commitments

$$\Phi_\sigma = \{\varphi_1, \ldots, \varphi_m\}$$

together with a linking theory $\Delta_\sigma^\to$ consisting solely of implication constraints.

The combined system $\Phi_\sigma \cup \Delta_\sigma^\to$ is satisfiable. Its joint satisfiability explicitly encodes the obstruction to the existence of any classical total valuation satisfying $\sigma$. In particular, the implication-only structure of $\Delta_\sigma^\to$ preserves local satisfiability of the commitments while preventing the collapse of cyclic self-reference into definitional inconsistency.

In the cyclic case, $\Phi_\sigma \cup \Delta_\sigma^\to$ is designed to remain satisfiable even when $1^n \notin C(\sigma)$; the preserved content is the *exclusion* of the total pattern (obstruction), not joint unsatisfiability.

*Remark.* In cyclic cases, the correspondence asserted here is obstruction-preserving rather than value-preserving: the joint satisfiability condition does not resolve $\sigma$ but faithfully represents the impossibility of a classical total valuation. The role of the implication-only linking theory is not to recover a global truth valuation, but to preserve the structural obstruction induced by cyclic self-reference in a consistency-preserving manner.

*Proof of Theorem 3.11 & 3.12*

Let $\sigma$ be a liar-type sentence in the self-referential fragment specified in Section 3.2. By definition, $\sigma$ is obtained via diagonalization and contains a truth predicate $T(\cdot)$ (or equivalent encoding) applied to its own Gödel code.

**Step 1:** Unfolding Self-Reference

Let $\mathcal{U}(\sigma)$ denote the finite unfolding of $\sigma$'s truth conditions obtained by replacing each occurrence of $T(\ulcorner \tau \urcorner)$ with a fresh propositional symbol $p_\tau$. Because $\sigma$ is finitely presented, $\mathcal{U}(\sigma)$ yields a finite set of fresh symbols $P_\sigma = \{p_1, \ldots, p_n\}$.

Each $p_i$ corresponds to an atomic semantic commitment induced by $\sigma$.

**Step 2: Linking theory construction**

From the finite unfolding $U(\sigma)$ we extract formulas $\theta_1, \ldots, \theta_n$ governing the intended semantic dependencies of fresh sentence symbols $p_1, \ldots, p_n$. Define the dependency graph $G_\sigma$ on vertices $\{1, \ldots, n\}$ by an edge

$$j \to i \text{ if and only if } p_j \text{ occurs in } \theta_i.$$

We define $\Delta_\sigma$ by cases.

**(Acyclic case.)** If $G_\sigma$ is acyclic, set

$$\Delta_\sigma = \{ p_i \leftrightarrow \theta_i \mid i = 1, \ldots, n \}.$$

By Lemma 3.11a, $\Delta_\sigma$ is classically consistent and admits expansions over every $M \in \mathcal{M}$.

**(Cyclic case.)** If $G_\sigma$ contains a directed cycle, biconditional "definitions" may impose inconsistent fixed-point constraints. In this case we use the cycle-safe linking theory

$$\Delta_\sigma^{\to} = \{ p_i \to \theta_i \mid i = 1, \ldots, n \}.$$

This theory is always classically consistent (Lemma 3.11c above). All semantic incompatibility is then captured at the level of admissible commitment patterns (the constraint set $C(\sigma)$) and the joint behavior of the family $I(\sigma)$, rather than being forced into $\Delta_\sigma$ itself.

**Step 3:** Definition of the INF Set

Let

$$I(\sigma) := \{ \varphi_i := p_i \mid i = 1, \ldots, n \}.$$

Each $\varphi_i$ is a non-self-referential sentence of the expanded language $L'$.

**Step 4:** Local Satisfiability

Fix any $M \in \mathcal{M}$. Because each $\theta_i$ is classically consistent in isolation, there exists an assignment

$$v_i : p_i \mapsto \{0, 1\}$$

such that $v_i(p_i) = 1$ and $v_i \models p_i \leftrightarrow \theta_i$.

Thus, for each $i$, there exists an expansion $M'_i \models \Delta_\sigma \wedge \varphi_i$. Hence each $\varphi_i$ is locally satisfiable under $\Delta_\sigma$.

**Step 5:** Global Obstruction

Suppose, for contradiction, that there exists an expansion $M' \models \Delta_\sigma \wedge \bigwedge_{i=1}^{n} \varphi_i$.

Then the interpretation of the $p_i$ induced by $M'$ yields a total assignment satisfying all linking equivalences.

Since the dependency graph of $\sigma$ is acyclic, the unfolded commitments $p_1, \ldots, p_n$ can be ordered topologically so that the defining condition for $p_i$ depends only on $p_j$ with $j < i$.

For any expansion $M' \models \Delta_\sigma$, define $v_i = 1$ if $M' \models \varphi_i$ and $v_i = 0$ otherwise. Because $\Delta_\sigma$ consists of definitional equivalences $p_i \leftrightarrow \theta_i(p_1, \ldots, p_{i-1})$, a straightforward induction on $i$ shows that the valuation $v$ uniquely determines the truth values of all unfolded components appearing in $U(\sigma)$.

Hence $\sigma$ admits a classical model if and only if $\Phi_\sigma \cup \Delta_\sigma$ is satisfiable. Thus, we prove that by construction of $\Delta_\sigma$, this assignment induces a classical truth valuation for all unfolded components of $\sigma$, and hence a classical global truth assignment for $\sigma$ itself.

This contradicts the unfolding constraint system: by definition, the Boolean dependency constraints induced by $U(\sigma)$ rule out the total pattern $1^n$ (equivalently, $1^n \notin C(\sigma)$). Hence no expansion $M'$ satisfying $\Delta_\sigma$ can satisfy $\bigwedge_{i=1}^{n} \varphi_i$. Therefore,

$$M' \not\models \bigwedge_{i=1}^{n} \varphi_i$$

for all expansions $M' \models \Delta_\sigma$.

This induction shows that the induced valuation $v$ uniquely determines the truth values of all unfolded components appearing in $U(\sigma)$ (i.e., all definitional right-hand sides $\theta_i$). In particular, $M' \models \Delta_\sigma$ implies that $v$ satisfies the Boolean dependency constraints extracted from $U(\sigma)$.

**Step 6:** Faithfulness

Conversely, suppose $\sigma$ admits a classical global truth assignment in $M$. Then the induced valuation of its unfolded components yields an assignment to $P_\sigma$ satisfying all equivalences in $\Delta_\sigma$. Hence there exists an expansion $M' \models \Delta_\sigma \wedge \bigwedge I(\sigma)$.

Thus, $\sigma$ admits a classical truth assignment **iff** $I(\sigma)$ is jointly satisfiable under $\Delta_\sigma$.

Unlike the acyclic case, the implication-only linking theory does not determine a total classical truth valuation for $\sigma$. Instead, it preserves local realizability of commitments together with the obstruction to joint satisfaction induced by cyclic self-reference.

**Conclusion**

All three conditions of Definition 3.10 are satisfied. Therefore, the INF construction is constraint-preserving and correct. ∎

*Remark 3.1.1a (Finite unfolding assumption)*

Our correctness proof assumes that the unfolding operator $\mathcal{U}(\sigma)$ yields a finite dependency graph on fresh sentence symbols. This holds for the *liar-type fragment considered here*, namely sentences whose truth conditions, after expanding the defining equations for the truth predicate $T$ (or the corresponding Gödel-coded truth schema), generate a finite system of Boolean constraints. This covers the standard liar, strengthened liar, and finite mutual-reference cycles. More general forms of semantic circularity that induce infinite dependency chains fall outside the present scope and are left for future work. Throughout this paper, the INF transformation is defined only for sentences $\sigma$ whose truth-predicate expansion terminates after finitely many steps. Concretely, $\sigma$ is said to admit a *finite unfolding* if there exists $n < \infty$ such that iterative expansion of all occurrences of the truth predicate $T(\ulcorner \cdot \urcorner)$ yields a finite family of propositional commitments $\Phi_\sigma = \{\varphi_1, \dots, \varphi_n\}$ together with a finite system of Boolean dependency constraints among them.

All correctness results in Section 3 are conditional on $\sigma$ belonging to this finite unfolding class. This class includes standard liar sentences, strengthened liars, and finite mutual-reference cycles, for which the unfolding depth and number of induced commitments are bounded by the syntactic nesting depth of the truth predicate. In each of these cases, the dependency graph induced by the unfolding has finitely many vertices and edges.

The framework does **not** currently apply to sentences whose unfolding generates an infinite constraint system. This includes, in particular:

- infinitary paradoxes such as Yablo-type constructions,
- sentences whose truth-predicate expansion produces unbounded chains of new commitments,
- self-referential sentences involving quantification over truth predicates, and
- constructions whose unfolding does not stabilize after finitely many steps.

These cases fall outside the scope of the present paper.

Truth-teller sentences and Curry-type sentences may or may not admit finite unfoldings depending on their syntactic form. Simple truth-tellers with bounded self-reference admit finite unfoldings but may generate multiple admissible commitment patterns; such multiplicity is compatible with the INF framework. Curry sentences with nested or unbounded conditional structure may fail to admit finite unfoldings and are therefore not covered in general.

**Example 3.12 (The Liar Sentence; cycle-safe INF with consistent linking theory).**
Let $\sigma$ be the liar sentence, informally "$\sigma$ is not true." In the INF construction we introduce two fresh sentence symbols $p_1, p_2$, intended to represent two incompatible semantic commitments extracted from the unfolding of $\sigma$:

- $p_1$: "$\sigma$ is true,"
- $p_2$: "$\sigma$ is not true" (equivalently: "$\sigma$ is false" in the classical fragment).

Because the unfolding dependency is cyclic, we use a **cycle-safe** linking theory consisting only of implications:

$$\vec{\Delta}_\sigma := \{ p_1 \to \neg p_2,\ p_2 \to \neg p_1 \}.$$

This theory is classically consistent (for example, the assignment $p_1 = 0, p_2 = 0$ satisfies all implications).

Define the INF family by

$$I(\sigma) := \{\varphi_1, \varphi_2\} \text{ where } \varphi_1 := p_1,\ \varphi_2 := p_2.$$

Then:

**(i) Local satisfiability.** Each $\varphi_i$ is satisfiable under $\vec{\Delta}_\sigma$. $p_1 = 1, p_2 = 0$ satisfies $\vec{\Delta}_\sigma$ and makes $\varphi_1$ true; and $p_1 = 0, p_2 = 1$ satisfies $\vec{\Delta}_\sigma$ and makes $\varphi_2$ true.

**(ii) No joint satisfiability.** No expansion $M' \models \vec{\Delta}_\sigma$ satisfies $\varphi_1 \wedge \varphi_2$. $p_1 \wedge p_2$ contradicts $p_1 \to \neg p_2$ (and also $p_2 \to \neg p_1$).

Thus, the liar obstruction is represented structurally as an **incongruent pair of locally admissible commitments**: each commitment is individually realizable, but they cannot be realized simultaneously under the consistent linking theory $\vec{\Delta}_\sigma$.

**Note.** In cyclic cases, using biconditionals (e.g., $p_1 \leftrightarrow \neg p_2$) can force $\Delta_\sigma$ itself to become inconsistent. The implication-only form $\vec{\Delta}_\sigma$ avoids that and keeps all inconsistency at the level of the joint family $I(\sigma)$.

*Remark 3.13*

The INF construction does not eliminate paradox by weakening semantics or stratifying truth. Instead, it **relocates self-reference into structured incompatibility** among non-self-referential sentences. The paradox is preserved as an explicit semantic constraint rather than as a syntactic anomaly.

*Example 3.13 (Curry Sentence)*

Let $\sigma$ be the Curry sentence: "If $\sigma$ is true, then $\bot$."
The unfolding dependency graph again contains a directed cycle.

Introduce a fresh sentence symbol $p_1$ representing the semantic commitment "$\sigma$ is true," and define the linking theory

$$\vec{\Delta}_\sigma = \{ p_1 \to \bot \}.$$

This theory is classically consistent.

Define the INF family

$$I(\sigma) = \{\varphi_1\}, \text{ where } \varphi_1 := p_1.$$

**Local satisfiability.**
There exists an expansion $M' \models \Delta_\sigma^\rightarrow$ with $M' \models \neg\varphi_1$, i.e., $p_1 = 0$.

**Global obstruction.**
There is no expansion $M' \models \Delta_\sigma^\rightarrow$ with $M' \models \varphi_1$, since $p_1 = 1$ implies $\bot$.

Thus $I(\sigma)$ is incongruent under $\Delta_\sigma^\rightarrow$, and the impossibility of satisfying $\varphi_1$ corresponds exactly to the semantic obstruction exhibited by the Curry sentence.

### 3.7 Discussion

Incongruent normal form separates **local coherence** from **global compatibility**. Classical semantics is preserved at the level of individual sentences, while inconsistency is isolated at the level of joint enforcement.

This shift allows self-referential paradoxes to be analyzed as structured semantic incompatibilities rather than failures of truth assignment.

*Remark (Edge cases)*

The unfolding constraint system may contain redundant equations or trivial self-reference that collapses to a tautology or contradiction at the level of Boolean constraints. Such cases are handled uniformly by the definition of $C(\sigma)$: redundancy does not affect the solution set, and trivialities correspond to $C(\sigma) = \{0,1\}^n$ or $C(\sigma) = \emptyset$ as appropriate.

### 3.8 Transition

In Section 4, we use incongruent sets to prove the main result of the paper: **semantically complete systems admit no informative sentences**, and therefore cannot support knowledge. Incompleteness emerges not as a limitation but as a structural requirement.

## 4 KNOWLEDGE REQUIRES SEMANTIC INCOMPLETENESS

In this section we analyze the relationship between semantic completeness and informativeness. Using only the structural notions developed in Sections 2 and 3, we show that a semantically complete model class admits no informative sentences.

Here "informativeness" is understood in a minimal, model-theoretic sense: a sentence is informative if it distinguishes between admissible semantic states or models. This notion concerns discriminative semantic power and does not presuppose any epistemic agent, proof system, or theory of justification. Accordingly, our results should be read as statements about the structure of semantic model classes, rather than about knowledge in an epistemological sense.

With this convention in place, the results below formalize a basic but important tension: enforcing global semantic determinacy collapses all contingent distinctions among models. This observation motivates the structural role of incongruence studied later in the paper.

## 4.1 Knowledge as Semantic Differentiation

Recall that a sentence is informative in a model class $\mathcal{M}$ if it distinguishes among admissible models. This notion captures the minimal logical content required for knowledge: without differentiation, no information is gained.

We first restate the key observation.

*Lemma 4.1 (Completeness Eliminates Informativeness)*

If $\mathcal{M}$ is semantically complete over a fragment $F$, then no sentence $\varphi \in F$ is informative in $\mathcal{M}$.

**Proof.**
By semantic completeness, for every $\varphi \in F$,

$$[[\varphi]]_\mathcal{M} \in \{\emptyset, \mathcal{M}\}.$$

Hence no $\varphi$ satisfies $\emptyset \neq [[\varphi]]_\mathcal{M} \neq \mathcal{M}$.

Therefore no informative sentences exist. ∎

This establishes that semantic completeness collapses all semantic degrees of freedom.

## 4.2 Incongruence as a Necessary Condition for Knowledge

We now show that incongruent sets provide the minimal structure required for informativeness.

*Lemma 4.2 (Incongruence Implies Semantic Differentiation)*

If $I = \{\varphi_1, \ldots, \varphi_n\}$ is an incongruent set in $\mathcal{M}$, then each $\varphi_i$ is informative in $\mathcal{M}$.

**Proof.**

By local satisfiability, $[[\varphi_i]]_\mathcal{M} \neq \emptyset$.

By global incompatibility, there exists at least one $\varphi_j$ such that $[[\varphi_i \wedge \varphi_j]]_\mathcal{M} = \emptyset$.

Hence $[[\varphi_i]]_\mathcal{M} \neq \mathcal{M}$.

Therefore each $\varphi_i$ is informative. ∎

Incongruence thus guarantees the existence of semantic alternatives.

## 4.3 Foundational Observation (Semantic Completeness Collapses Variability)

We now state the foundational observation.

*Observation 4.3 (Knowledge Requires Semantic Incompleteness)*

If a semantic system admits knowledge (i.e., informative sentences), then it is semantically incomplete.

Equivalently, any semantically complete system admits no knowledge.

**Proof (by definition).**
Assume the system admits knowledge. Then there exists an informative sentence $\varphi$. By Definition 2.2,

$$\emptyset \neq [[\varphi]]_{\mathcal{M}} \neq \mathcal{M}.$$

This contradicts semantic completeness.
Therefore, any system supporting knowledge must be semantically incomplete. ∎

This proposition formalizes the central thesis:

**Incompleteness is not a limitation on knowledge; it is a necessary condition for it.**

*Observation 4.3 is included as a structural observation; the substantive results begin with the quantitative bounds of Section 5. It is included to make explicit a structural tension already implicit in the definitions, rather than as an independent impossibility theorem.*

*Remark (Semantic vs. syntactic completeness).*
Semantic completeness, as used here, is a property of a model class: every sentence in a given fragment has either universal or empty truth set. This notion is distinct from syntactic completeness of a theory (decidability of sentences by axioms), and from Gödel incompleteness, which concerns provability limits in recursively axiomatized systems. The present observation concerns semantic contingency, not provability. Any comparison to Gödel's theorem is intended only at the level of structural analogy, not technical correspondence.

### 4.4 Tradeoff of Determinacy Across Incongruent Alternatives

Incongruent sets also impose a structural tradeoff.

*Proposition 4.4 (Determinacy Tradeoff)*

Let $I = \{\varphi_1, \ldots, \varphi_n\}$ be incongruent in $\mathcal{M}$.

For any refinement $\mathcal{M}' \subseteq \mathcal{M}$ such that $\mathcal{M}' \subseteq [[\varphi_k]]_{\mathcal{M}}$, there exists $j \neq k$ such that $[[\varphi_j]]_{\mathcal{M}'} = \emptyset$.

**Proof.**

Since $I$ is not jointly satisfiable, no model satisfies all $\varphi_i$.

Restricting $\mathcal{M}$ to enforce $\varphi_k$ necessarily excludes at least one other $\varphi_j$. ∎

Semantic determinacy along one-dimension forces collapse along another.

## 4.5 Paradox as a Structural Boundary

We now explain why paradoxes arise naturally in expressive systems.

*Corollary 4.5 (Paradox Pressure)*

In any expressive semantic system that:

1. supports incongruent sets, and
2. allows semantic assertions about its own sentences,

attempts to enforce global semantic completeness necessarily result in contradiction, loss of informativeness, or adoption of non-classical semantics.

**Explanation.**
By Theorem 4.3, semantic completeness eliminates knowledge.
By Section 3, self-referential sentences encode incongruent semantic structures.
Compressing such structures into a single globally complete description forces contradiction or collapse.

Thus, paradox marks the boundary of semantic collapse.

## 4.6 Incompleteness Generates Finite Incongruence

We now show that semantic incompleteness of an axiomatic theory can be understood structurally as the presence of *latent incongruence* among its admissible extensions. This result does not rely on self-reference or paradox and applies to ordinary first-order theories.

Let $T$ be a consistent first-order theory with model class $\mathcal{M}_T$. Recall that $T$ is *complete* if for every sentence $\varphi$, either $T \vdash \varphi$ or $T \vdash \neg\varphi$, and *incomplete* otherwise.

*Proposition 4.6 (Extension-Induced Incongruence).*
Let $T$ be a consistent first-order theory over a language $L$, and let $\mathcal{M}_T$ denote its class of $L$-models. If $T$ is incomplete, then there exists a finite family of $L$-sentences

$$\Phi = \{\varphi_1, \ldots, \varphi_n\}$$

such that:

1. (**Local satisfiability**) For each $i$, there exists $M_i \in \mathcal{M}_T$ with $M_i \vDash \varphi_i$ (equivalently, $T \cup \{\varphi_i\}$ is satisfiable);
2. (**Global obstruction**) $T \cup \Phi$ is unsatisfiable (equivalently, no $M \in \mathcal{M}_T$ satisfies all $\varphi_i$ simultaneously).
   Consequently, $\Phi$ is incongruent relative to $\mathcal{M}_T$ in the sense of Definition 2.6.

**Proof.**
Since $T$ is incomplete, there exists a sentence $\psi \in \text{Sent}(L)$ such that

$$T \nvdash \psi \text{ and } T \nvdash \neg\psi.$$

By soundness and completeness of first-order logic, this implies that there exist models

$$M^+, M^- \in \mathcal{M}_T$$

such that

$$M^+ \models \psi \text{ and } M^- \models \neg\psi.$$

Define the finite set

$$I = \{\psi, \neg\psi\}.$$

We verify that $I$ is incongruent in $\mathcal{M}_T$:

1. **Local satisfiability.**
   By construction,

   $$[[\psi]]_{\mathcal{M}_T} \neq \emptyset \text{ (witnessed by } M^+),$$
   $$[[\neg\psi]]_{\mathcal{M}_T} \neq \emptyset \text{ (witnessed by } M^-).$$

2. **No joint satisfiability.**
   For any structure $M$,

   $$M \models \psi \wedge \neg\psi$$

   is impossible. Hence

   $$[[\psi \wedge \neg\psi]]_{\mathcal{M}_T} = \emptyset.$$

Thus $I$ satisfies both conditions of Definition 2.6 and is incongruent in $\mathcal{M}_T$. ∎

*Remark 4.6.1 (Beyond Binary Splits)*

While Remark 4.6 exhibits the minimal incongruent set arising from incompleteness, more structured incongruent families can be obtained by selecting sentences from incompatible complete extensions of $T$ and applying compactness to their joint constraints. In such cases, no element of the incongruent family is the negation of another, and the incompatibility reflects genuine multi-way semantic obstruction rather than a binary choice.

*Proposition 4.6b (Genuine Multi-way Incongruence Without Negations)*

There exist consistent incomplete theories $T$ for which, for arbitrarily large $n$, one can construct finite incongruent sets

$$I = \{\varphi_1, \ldots, \varphi_n\} \subseteq \text{Sent}(L)$$

in $\mathcal{M}_T$ such that **no $\varphi_i$ is logically equivalent to $\neg\varphi_j$** for any $i \neq j$.

**Proof.**

Fix $n \geq 2$. Let $L$ be a first-order language containing unary predicate symbols $P_1, \ldots, P_n$ and a constant symbol $c$. Let $T$ be the $L$-theory consisting of the single axiom

$$\forall x \bigwedge_{1 \leq i < j \leq n} \neg(P_i(x) \wedge P_j(x)),$$

i.e., for every element $x$, at most one of $P_1(x), \ldots, P_n(x)$ holds. This theory $T$ is consistent (e.g., take a one-element model with all $P_i$ false) and incomplete (e.g., it does not decide $P_1(c)$).

For each $i \in \{1, \ldots, n\}$, define the sentence

$$\varphi_i := P_i(c).$$

Let $I = \{\varphi_1, \ldots, \varphi_n\}$.

We verify that $I$ is incongruent in $\mathcal{M}_T$ (Definition 2.6):

1. **Local satisfiability.** For each $i$, there exists $M_i \models T$ with $M_i \models P_i(c)$: take a one-element structure in which $c$ denotes that element and interpret $P_i$ as true on it, while all other $P_j$ are false. Hence

$$[[\varphi_i]]_{\mathcal{M}_T} \neq \emptyset \text{ for all } i.$$

2. **No joint satisfiability.** In any $M \models T$, the axiom implies that for the element $c^M$, at most one of $P_1(c), \ldots, P_n(c)$ can hold. Therefore no $M \models T$ can satisfy $\bigwedge_{i=1}^{n} P_i(c)$, and thus

$$[[\bigwedge_{i=1}^{n} \varphi_i]]_{\mathcal{M}_T} = \emptyset.$$

Hence $I$ is incongruent in $\mathcal{M}_T$.

Finally, observe that for $i \neq j$, $\varphi_i$ is not logically equivalent to $\neg\varphi_j$ over $\mathcal{M}_T$: indeed, there are models of $T$ in which $P_i(c)$ and $P_j(c)$ are both false (e.g., the model with all $P_k$ false), so $\neg\varphi_j$ can hold while $\varphi_i$ fails. Thus, the family is genuinely multi-way and not reducible to binary negation pairs. ∎

*While the existence of undecidable sentences is classical, Proposition 4.6b shows that incompleteness induces genuinely multi-way incongruence not reducible to negation pairs.*

*Remark 4.6.2 (Connection to completions)*

The theory $T$ admits distinct complete extensions $T_i := T \cup \{P_i(c)\} \cup \{\neg P_j(c) : j \neq i\}$, each consistent and complete in the fragment generated by the $P_j(c)$. Proposition 4.6b shows that selecting sentences from such mutually incompatible completions yields finite incongruent families in which incompatibility is structural rather than a single sentence/negation split.

*Remark 4.6.3 (Beyond Independence)*

This phenomenon is strictly stronger than the existence of an undecidable sentence. Independence concerns a single sentence whose truth value is undetermined by $T$, whereas incongruence concerns *families* of sentences whose individual realizability masks a joint semantic obstruction. In this sense, incompleteness manifests not merely as indeterminacy but as structured incompatibility among admissible extensions.

*Remark 4.6.4 (Structural Interpretation)*

From this perspective, incompleteness prevents the collapse of latent incongruence into outright contradiction. A complete theory eliminates all such families by forcing global determinacy, but at the cost of semantic differentiation. Incomplete theories, by contrast, support nontrivial incongruent families, which enable semantic variability without inconsistency.

*The preceding results show that incongruence is not confined to semantic paradox or self-reference, but arises generically from incompleteness itself. Any incomplete theory necessarily supports finite families of locally realizable but jointly incompatible semantic commitments, ranging from minimal binary splits to genuinely multi-way obstruction. These incongruent families represent the semantic degrees of freedom preserved by incompleteness. In Section 5 we show that maintaining such degrees of freedom carries a quantitative cost: in finite semantic-state models, incongruence constrains how semantic variance and determinacy can be distributed, yielding uncertainty-style bounds on semantic representation.*

## 4.7  Interpretation

The results of this section admit a simple structural interpretation. Incongruence marks the presence of genuine semantic degrees of freedom: distinct commitments that are each locally realizable but cannot be realized simultaneously. Informativeness, in the model-theoretic sense used here, consists in differentiating among these alternatives by restricting the class of admissible models without collapsing it to a single outcome.

Semantic completeness removes such degrees of freedom by enforcing global determinacy: every sentence is either satisfied by all admissible models or by none. While this guarantees consistency, it also eliminates all contingent distinctions and thus precludes informativeness. Semantic incompleteness, by contrast, preserves structured variability among admissible models, allowing informative distinctions to coexist with classical semantics.

From this perspective, paradox does not arise from incongruence itself, but from attempts to impose global determinacy in expressive systems where incompatible commitments are unavoidable. The transformation to incongruent normal form makes this obstruction explicit by separating local semantic coherence from global incompatibility.

In this limited sense, incompleteness is necessary for informativeness: contingent sentences require model classes with multiple admissible realizations. This claim is purely model-theoretic. No assertion is made about human cognition, epistemic justification, or broader philosophical theories of knowledge.

Finally, we note that the existence of incongruent families in incomplete theories (Propositions 4.6–4.6b) does not rely on any finiteness assumption. By contrast, the quantitative variance bounds developed in Section 5 apply only to finite semantic-state spaces and should be understood as analytically tractable approximations rather than universal constraints.

# 5 SEMANTIC ENERGY AND QUANTITATIVE SEMANTIC UNCERTAINTY

This section makes the value of semantic incompleteness quantitative. We introduce a spectral/Dirichlet notion of **semantic energy** and prove that semantic determinacy, informational content, and spectral simplicity cannot be simultaneously optimized. We also give a matrix formulation for families of sentences.

*Relationship to Section 4.*
The results of Section 4 are purely model-theoretic and apply to arbitrary (possibly infinite) classes of structures. By contrast, the quantitative bounds developed in this section require a finite semantic-state space equipped with a reversible local-variation operator. Accordingly, the variance–energy inequalities derived below do not directly constrain infinite model classes; they apply to finite propositional or finite-state approximations arising, for example, from unfolded commitment families.

## 5.0 General finite semantic state spaces

The results of Sections 2–4 are model-theoretic and apply to arbitrary classes of structures. In this section we study a restricted—but analytically tractable—semantic-state setting that permits quantitative bounds.

We interpret the model class $\mathcal{M}$ as a space of admissible semantic states, and each sentence $\varphi$ as an indicator function $f_\varphi: \mathcal{M} \to \{0,1\}$ of its truth set. To obtain explicit uncertainty-style inequalities, we assume $\mathcal{M}$ is finite and is equipped with a notion of reversible "local semantic variation." Concretely, we model local variation by a reversible Markov kernel.

In settings arising from the INF transformation, the semantic state space admits a concrete finite representation. Given an INF-transformed sentence $\sigma$ with commitment family $\Phi_\sigma = \{\varphi_1, \ldots, \varphi_m\}$, semantic states may be identified with admissible joint truth assignments to these commitments, subject to the associated linking constraints $\Delta_\sigma$. This yields a finite semantic state space $\Omega \subseteq \{0,1\}^m$, where each coordinate records the truth value of a commitment. In the absence of additional structural information, $\Omega$ is equipped with the uniform probability measure $\pi$. Local semantic variation may then be modeled by single-coordinate perturbations consistent with $\Omega$, inducing a reversible Markov kernel on $\Omega$. When $\Omega = \{0,1\}^m$, this construction coincides with the Boolean hypercube equipped with its standard edge-flip dynamics.

Formally, let $\mathcal{M}$ be a finite set, let $\pi$ be a probability measure on $\mathcal{M}$ (uniform unless stated otherwise), and let $P(x,y)$ be a Markov kernel on $\mathcal{M}$ that is reversible with respect to $\pi$ (i.e., $\pi(x)P(x,y) = \pi(y)P(y,x)$). The triple $(\mathcal{M}, P, \pi)$ induces a Dirichlet form that measures local sensitivity of a function $g: \mathcal{M} \to \mathbb{R}$ via

$$\mathcal{E}(g,g) := \frac{1}{2} \sum_{x,y \in \mathcal{M}} \pi(x) \, P(x,y) \, (g(x) - g(y))^2.$$

This framework subsumes the Boolean cube with single-bit flips as the canonical special case $\mathcal{M} = \{0,1\}^m$ with $\pi$ uniform and $P$ the random single-coordinate flip kernel; in that case, $\mathcal{E}(g,g)$ coincides (up to the standard normalization) with total influence / boundary sensitivity.

Throughout Section 5 we assume $(\mathcal{M}, P, \pi)$ satisfies a Poincaré inequality with constant $\lambda > 0$, meaning that for every $g: \mathcal{M} \to \mathbb{R}$,

$$\mathrm{Var}_\pi(g) \leq \frac{1}{\lambda}\mathcal{E}(g,g).$$

All expectations and variances in Section 5 are taken with respect to $\pi$. When we require closed-form constants and an explicit Fourier basis, we specialize to the Boolean cube (the setting in which the sharp entropy–influence tradeoff is standard in the analysis of Boolean functions).

**Scope of the quantitative analysis.**
The uncertainty and variance bounds derived in Section 5 apply to finite semantic state spaces, such as those arising from finite propositional fragments or finite commitment families produced by the INF transformation. We do not claim that these bounds directly extend to arbitrary infinite first-order model classes. Rather, they quantify structural tradeoffs within finite semantic approximations, which serve as analytically tractable surrogates for more complex semantic domains.

### 5.1 Semantic State Space and Indicators

Let $G = \mathbb{Z}_2^m$ be the space of admissible semantic states ("worlds"), equipped with the uniform measure. A sentence $\varphi$ is represented by its truth indicator

$$f_\varphi : G \to \{0,1\}, \quad f_\varphi(x) = \mathbf{1}_{\{x \models \varphi\}}.$$

Let

$$p_\varphi := \mathbb{E}[f_\varphi] = \Pr[f_\varphi = 1].$$

Define the **semantic knowledge content** of $\varphi$ as the entropy of its truth value:

$$K(\varphi) := H(f_\varphi) = h(p_\varphi),$$

where $h$ is the binary entropy (base 2).

The choice of total influence as semantic energy is motivated by its interpretation as *boundary sensitivity*. Influence measures how often the truth value of a sentence changes under minimal perturbations of the underlying semantic state. A sentence with low influence is stable under semantic variation and thus semantically rigid; a sentence with high influence depends delicately on many semantic degrees of freedom.

This notion aligns naturally with semantic complexity: informative sentences must distinguish among many admissible models, which necessarily requires sensitivity to variations in semantic state. Total influence provides the minimal quantity that (i) captures this sensitivity, (ii) composes additively across independent semantic dimensions, and (iii) supports sharp entropy and uncertainty inequalities. Alternative complexity measures (e.g., circuit size or description length) capture different aspects of computation but do not yield comparable analytic bounds on semantic entropy.

### 5.2 Fourier Expansion and Semantic Energy

Work with the $\{-1, +1\}$ encoding

$$g_\varphi(x) := 2f_\varphi(x) - 1 \in \{-1, +1\}.$$

Let $\hat{g}_\varphi(S)$ denote the Fourier coefficients of $g_\varphi$ on $\mathbb{Z}_2^m$. Then Parseval gives

$$\sum_{S \subseteq [m]} \hat{g}_\varphi(S)^2 = \mathbb{E}[g_\varphi^2] = 1.$$

*Definition 5.1 (Semantic Energy / Total Influence)*

Define the **semantic energy** of $\varphi$ as the total influence (Dirichlet energy)

$$E(\varphi) := \sum_{i=1}^{m} \text{Inf}_i(g_\varphi), \text{ where } \text{Inf}_i(g) := \Pr_{x \sim G}[g(x) \neq g(x^{\oplus i})],$$

and $x^{\oplus i}$ denotes $x$ with bit $i$ flipped.

Equivalently (standard Fourier identity on the cube),

$$E(\varphi) = \sum_{S \neq \emptyset} |S| \; \hat{g}_\varphi(S)^2.$$

Thus $E(\varphi)$ is a bona-fide Fourier-weighted "energy": it measures how much spectral mass lies on higher-order modes, weighted by their order.

Intuitively, semantic energy measures the sensitivity of a sentence's truth value to local semantic perturbations: a sentence has low semantic energy if its truth is stable under small changes to the underlying semantic state, and high energy if its truth value frequently flips under such perturbations. In logical terms, this corresponds to how finely a sentence "cuts" the semantic state space, rather than to syntactic complexity alone.

### 5.3 Semantic Uncertainty Inequality

We now connect knowledge content $K(\varphi) = h(p_\varphi)$ to energy $E(\varphi)$.

*Lemma 5.2 (Poincaré Inequality on the Cube)*

For any $g: G \to \mathbb{R}$,

$$\text{Var}(g) \leq \sum_{i=1}^{m} \mathbb{E}[(g(x) - g(x^{\oplus i}))^2]/4.$$

For $g_\varphi \in \{-1, +1\}$, this reduces to

$$\text{Var}(g_\varphi) \leq E(\varphi).$$

Since $\text{Var}(g_\varphi) = 1 - (\mathbb{E}[g_\varphi])^2$ and $\mathbb{E}[g_\varphi] = 2p_\varphi - 1$, we have

$$1 - (2p_\varphi - 1)^2 \le E(\varphi).$$

Equivalently,

$$4p_\varphi(1 - p_\varphi) \le E(\varphi).$$

*Proof.*

This is the standard Poincaré inequality on the Boolean cube; see O'Donnell, *Analysis of Boolean Functions*, Theorem 9.21. We include the short algebraic specialization used here.

Fix a coordinate $i$. Since $g$ takes values in $\{-1, +1\}$, the squared difference across the $i$-th edge satisfies

$$(g(x) - g(x^{\oplus i}))^2 = \begin{cases} 0, & \text{if } g(x) = g(x^{\oplus i}), \\ 4, & \text{if } g(x) \ne g(x^{\oplus i}). \end{cases}$$

Equivalently, $(g(x) - g(x^{\oplus i}))^2 = 4 \cdot \mathbf{1}\{g(x) \ne g(x^{\oplus i})\}$. Taking expectations gives

$$\frac{1}{4}\mathbb{E}[(g(x) - g(x^{\oplus i}))^2] = \Pr[g(x) \ne g(x^{\oplus i})] = \mathrm{Inf}_i(g),$$

the influence of coordinate $i$.

Applying the Poincaré inequality and substituting the above identity yields the explicit cancellation:

$$\mathrm{Var}(g) \le \frac{1}{4}\sum_{i=1}^{m} \mathbb{E}[(g(x) - g(x^{\oplus i}))^2] = \frac{1}{4}\sum_{i=1}^{m} 4\,\mathrm{Inf}_i(g) = \sum_{i=1}^{m} \mathrm{Inf}_i(g) = E(\varphi),$$

which proves the claim.

*Remark (sharpness).* The constant $1/4$ is the optimal Poincaré constant for the Boolean cube; equality is achieved for first-level Fourier characters (dictator functions), so the inequality is sharp. ∎

So energy lower-bounds variance (hence "semantic non-collapse"). Next we convert this into an entropy bound.

*Lemma 5.3 (Entropy Upper Bound via Minimal Mass)*

Let $q := \min(p, 1-p) \in [0, 1/2]$. Then

$$h(p) = h(q) \le q\log_2\left(\frac{e}{q}\right).$$

*Proof.* Standard bound $h(q) \le q\log(e/q)$ obtained from concavity of $\log$ and the inequality $-(1-q)\log(1-q) \le q$. ∎

Now combine Lemma 5.2 and the elementary inequality $q \leq 2p(1-p)$ for $q \in [0,1/2]$ (since $p(1-p) = q(1-q) \geq q/2 \Rightarrow q \leq 2p(1-p)$).

So $q \leq 2p(1-p) \leq E(\varphi)/2$.

*Theorem 5.4 (Semantic Uncertainty Bound: Energy ⇒ Entropy).*

Let $f: \{0,1\}^m \to \{0,1\}$ be a Boolean indicator function and let $E(f)$ denote its total influence with respect to the uniform measure. Let $K(f) = H(\mathbb{P}[f = 1])$ denote its binary entropy.

If $E(f) = 0$, then $K(f) = 0$.

If $0 < E(f) \leq 1$, then

$$K(f) \leq \min\{1, \frac{E(f)}{2} \log_2(2e/E(f))\}.$$

For $E(f) > 1$, the bound is trivial since $K(f) \leq 1$.

*Proof.*

**Case $E(\varphi) = 0$:** If $E(\varphi) = 0$, then all influences vanish, hence $g_\varphi$ is constant on the cube, so $p_\varphi \in \{0,1\}$ and $K(\varphi) = 0$.

**Case $E(\varphi) > 0$:** Let $q = \min(p_\varphi, 1 - p_\varphi)$. If $q = 0$, then $K(\varphi) = 0$ and the inequality holds. Otherwise $q \in (0, 1/2]$. By Lemma 5.2,

$$4p_\varphi(1 - p_\varphi) \leq E(\varphi) \Rightarrow 2p_\varphi(1 - p_\varphi) \leq E(\varphi)/2.$$

Since $p_\varphi(1 - p_\varphi) = q(1-q) \geq q/2$, we obtain $q \leq 2p_\varphi(1 - p_\varphi) \leq E(\varphi)/2$.
Applying Lemma 5.3 and monotonicity of $x \mapsto x \log_2(e/x)$ on $(0,1]$,

$$K(\varphi) = h(q) \leq q \log_2(e/q) \leq \frac{E(\varphi)}{2} \log_2(\frac{2e}{E(\varphi)}).$$

Also $h(p_\varphi) \leq 1$. Taking the minimum yields the claim. ∎

*This inequality is a direct instantiation of classical entropy–influence bounds for Boolean functions, here interpreted semantically via the indicator representation of truth.*

*Scope remark.* The entropy–energy bound is informative only in the low-energy regime, corresponding to predicates with limited boundary sensitivity. For highly complex predicates depending on many semantic dimensions, total influence typically exceeds this regime and the bound becomes vacuous. The result should therefore be read as a structural constraint on low-complexity semantic commitments rather than a universal limitation.

*Remark (when the bound is nontrivial):*

Write the right-hand side as $K(\varphi) \leq \min\{1, B(E(\varphi))\}$, where:

$$B(E) := \frac{E}{2} \log_2\left(\frac{2e}{E}\right).$$

Since $K(\varphi) \leq 1$ always, the inequality is informative exactly when $B(E(\varphi)) < 1$. The function $B(E)$ crosses 1 at a unique threshold $E^{\backslash *} \in (0, 2]$ defined by

$$\frac{E^{\backslash *}}{2} \log_2\left(\frac{2e}{E^{\backslash *}}\right) = 1$$

and numerically $E^{\backslash *} \approx 0.6551$ under the normalization in this section. Thus, for $E(\varphi) < E^{\backslash *}$ the theorem gives a genuine quantitative restriction $K(\varphi) \leq B(E(\varphi))$, while for $E(\varphi) \geq E^{\backslash *}$ it reduces to the trivial bound $K(\varphi) \leq 1$. In particular, the semantic uncertainty bound constrains the low-energy ("spectrally simple / low boundary sensitivity") regime and becomes vacuous once energy is large enough to permit full one-bit variability.

**Example sanity checks.**
• If $\varphi$ is constant, then $E(\varphi) = 0$ and $K(\varphi) = 0$, matching the bound.
• If $\varphi$ is maximally balanced with high boundary sensitivity, $K(\varphi)$ can approach 1, and the $\min{1, ...}$ form correctly reflects that the inequality is not meant to constrain this high-energy regime.

Note that $B(E)$ is continuous on $(0, \infty)$, tends to 0 as $E \to 0+$, and $B(1) = (1/2) \log 2(2e) > 1$, hence there is a unique crossing $E* \in (0, 1)$.

**What this gives us:** a real, quantitative law:

- if $E(\varphi)$ is forced small (spectrally simple / low boundary / low influence), then $K(\varphi)$ must be small (little knowledge content);
- to maintain $K(\varphi) \geq \kappa$, energy must be bounded away from zero (implicit lower bound).

The inequality in Theorem 5.4 follows from standard Poincaré and entropy inequalities on the Boolean cube. The contribution here is not the inequality itself, but its semantic interpretation: it shows that informativeness (entropy of truth) is quantitatively constrained by semantic sensitivity. This provides a precise formulation of a tradeoff that is implicit in semantic reasoning but rarely made explicit.

**Semantic Uncertainty Principle (Energy–Knowledge).**
Let $\varphi$ be any sentence represented on $G = \mathbb{Z}_2^m$ by $f_\varphi$ and $g_\varphi = 2f_\varphi - 1$. Let $E(\varphi) = \sum_i \mathrm{Inf}_i(g_\varphi) = \sum_{S \neq \emptyset} |S| \hat{g}_\varphi(S)^2$ be its semantic energy, and let $K(\varphi) = h(p_\varphi)$ be the entropy of its truth value across admissible states. Then

$$K(\varphi) \leq \min\left\{1, \frac{E(\varphi)}{2} \log_2\left(\frac{2e}{E(\varphi)}\right)\right\}.$$

In particular, **semantic completeness** ($p_\varphi \in \{0, 1\}$) forces $K(\varphi) = 0$, and **low semantic energy forces low knowledge content**. Attempts to make sentences globally determined while keeping them informative require non-negligible semantic energy.

## 5.4 Matrix Uncertainty for Many Sentences

Let $\Phi = \{\varphi_1, \ldots, \varphi_n\}$ be a family of sentences. Define the centered $\{-1, +1\}$ encodings

$$g_i := g_{\varphi_i}, \bar{g}_i := g_i - \mathbb{E}[g_i].$$

Define the **semantic covariance (Gram) matrix**

$$\Sigma \in \mathbb{R}^{n \times n}, \Sigma_{ij} := \mathbb{E}[\bar{g}_i \, \bar{g}_j].$$

Then $\mathrm{Tr}(\Sigma) = \sum_i \mathrm{Var}(g_i)$.

Define the **energy matrix**

$$\mathcal{E}_{ij} := \sum_{k=1}^{m} \mathbb{E}[\frac{(\bar{g}_i(x) - \bar{g}_i(x^{\oplus k}))(\bar{g}_j(x) - \bar{g}_j(x^{\oplus k}))}{4}].$$

This is the Dirichlet-form bilinearization; in particular $\mathcal{E}_{ii} = E(\varphi_i)$.

*Theorem 5.5 (Matrix Semantic Poincaré Inequality)*
$$\Sigma \preceq \mathcal{E} \text{(positive semidefinite order)}.$$

In particular,

$$\mathrm{Tr}(\Sigma) \leq \mathrm{Tr}(\mathcal{E}) = \sum_{i=1}^{n} E(\varphi_i).$$

*Proof.*

Let $\tilde{g}(x) = (\tilde{g}_1(x), \ldots, \tilde{g}_r(x))^\top$ denote the vector of centered observables, where $\tilde{g}_i(x) = g_i(x) - \mathbb{E}[g_i]$. Let $\Sigma \in \mathbb{R}^{r \times r}$ be the covariance matrix with entries $\Sigma_{ij} = \mathbb{E}[\tilde{g}_i(x)\tilde{g}_j(x)]$. Fix an arbitrary coefficient vector $a \in \mathbb{R}^r$ and define the scalar function

$$h_a(x) := \sum_{i=1}^{r} a_i \, \tilde{g}_i(x) = a^\top \tilde{g}(x).$$

**Step 1 (Left-hand side as a quadratic form).** Since each $\tilde{g}_i$ has mean zero, $h_a$ also has mean zero, hence $\mathrm{Var}(h_a) = \mathbb{E}[h_a(x)^2]$. Expanding gives

$$\mathrm{Var}(h_a) = \mathbb{E}\left[\left(\sum_i a_i \, \tilde{g}_i(x)\right)\left(\sum_j a_j \, \tilde{g}_j(x)\right)\right] = \sum_{i,j} a_i a_j \, \mathbb{E}[\tilde{g}_i(x)\tilde{g}_j(x)] = a^\top \Sigma a.$$

**Step 2 (Apply scalar Poincaré and expand the right-hand side).** By the scalar Poincaré inequality on the Boolean cube (see, e.g., O'Donnell, *Analysis of Boolean Functions*, Theorem 9.21; or Ledoux, *The Concentration of Measure Phenomenon*, Chapter 3 for the general Dirichlet-form formulation), we have

$$\mathrm{Var}(h_a) \leq \frac{1}{4} \sum_{k=1}^{m} \mathbb{E}[(h_a(x) - h_a(x^{\oplus k}))^2].$$

For each $k$, expand the difference:

$$h_a(x) - h_a(x^{\oplus k}) = \sum_{i=1}^{r} a_i \, (\tilde{g}_i(x) - \tilde{g}_i(x^{\oplus k})).$$

Squaring and taking expectation yields

$$\mathbb{E}[(h_a(x) - h_a(x^{\oplus k}))^2] = \mathbb{E}\left[\left(\sum_i a_i \, (\tilde{g}_i(x) - \tilde{g}_i(x^{\oplus k}))\right)\left(\sum_j a_j \, (\tilde{g}_j(x) - \tilde{g}_j(x^{\oplus k}))\right)\right]$$

$$= \sum_{i,j} a_i \, a_j \, \mathbb{E}\left[(\tilde{g}_i(x) - \tilde{g}_i(x^{\oplus k}))(\tilde{g}_j(x) - \tilde{g}_j(x^{\oplus k}))\right].$$

Collecting coefficients of $a_i a_j$ yields $\sum_{i,j} a_i a_j \mathcal{E}_{ij} = a^\top \mathcal{E} a$, where $\mathcal{E}_{ij} := \sum_k \frac{1}{4} \mathbb{E}[(\tilde{g}_i(x) - \tilde{g}_i(x^{\oplus k}))(\tilde{g}_j(x) - \tilde{g}_j(x^{\oplus k}))]$.

Summing over $k$ and including the factor $1/4$, define the matrix $E \in \mathbb{R}^{r \times r}$ entrywise by

$$E_{ij} := \frac{1}{4} \sum_{k=1}^{m} \mathbb{E}\left[(\tilde{g}_i(x) - \tilde{g}_i(x^{\oplus k}))(\tilde{g}_j(x) - \tilde{g}_j(x^{\oplus k}))\right].$$

With this definition, the right-hand side of Poincaré becomes exactly the quadratic form $a^\top E a$. Combining with Step 1 gives, for all $a \in \mathbb{R}^r$,

$$a^\top \Sigma a = \mathrm{Var}(h_a) \leq a^\top E a.$$

Equivalently, $E - \Sigma$ is positive semidefinite, i.e., $\Sigma \preceq E$.

**Step 3 (Verify that $E$ is positive semidefinite).** The definition of $E$ shows it is a sum of Gram matrices. Indeed, for fixed $k$, let

$$\Delta_k(x) := \tilde{g}(x) - \tilde{g}(x^{\oplus k}) \in \mathbb{R}^r.$$

Then $E = \frac{1}{4} \sum_{k=1}^{m} \mathbb{E}[\Delta_k(x) \Delta_k(x)^\top]$. For any $a$,

$$a^\top E a = \frac{1}{4} \sum_{k=1}^{m} \mathbb{E}[(a^\top \Delta_k(x))^2] \geq 0,$$

so $E \succeq 0$ as required. (Moreover, this is exactly the matrix form of the Dirichlet energy associated to the cube.)

**Step 4 (Trace inequality).** Since $\Sigma \preceq E$, standard matrix theory implies $\text{Tr}(\Sigma) \leq \text{Tr}(E)$. Concretely, $\Sigma \preceq E$ means $a^\top \Sigma a \leq a^\top E a$ for all $a$, hence all eigenvalues of $\Sigma$ are bounded above by the corresponding eigenvalues of $E$, and summing eigenvalues yields the trace inequality. This completes the proof. ∎

**Interpretation:** across a *set* of sentences, the total "variance budget" (hence the possibility of informativeness) is bounded by total energy. This yields a genuinely multi-dimensional uncertainty relation, formulated in matrix form.

If $\Phi$ is an **incongruent set**, joint satisfiability constraints impose additional structure on $\Sigma$ (e.g., certain correlations must be negative or zero), strengthening tradeoffs — we can state those as corollaries once we formalize the exact intersection constraints our "incongruence" enforces.

## 5.5 Incongruence Tightening: Energy Redistribution Under Incompatibility

**Structural source of variance suppression.**
The variance bounds in this section arise from bounded overlap of semantic commitments, quantified by the width $w(\Phi) = \text{ess sup}_\omega S_\Phi(\omega)$. Width directly limits the number of commitments that can be simultaneously realized in any semantic state and is the sole mechanism driving the inequalities below. Incongruence is not required for these bounds to hold; rather, incongruence is one structural condition that guarantees a nontrivial upper bound on width (specifically $w(\Phi) \leq n - 1$).

We now show that incongruence imposes additional quantitative constraints beyond those implied by semantic uncertainty alone. Incongruent sets force **redistribution of semantic energy across alternatives**, yielding stronger bounds on informativeness.

*Remark (Width vs. incongruence)*

The variance bounds in this section depend on the width parameter $w(\Phi)$, which measures the maximum number of commitments simultaneously realizable in a single semantic state. Incongruence is one canonical semantic mechanism that enforces bounded width (for example, $w(\Phi) \leq n - 1$ for incongruent families), but the inequalities themselves are width-based and apply to arbitrary families. We do not claim that incongruence alone yields stronger bounds than width-matched non-incongruent families; rather, incongruence constrains how total semantic variance can be redistributed across incompatible alternatives.

### 5.5.1 Incongruence as Orthogonality Constraint

Let $\Phi = \{\varphi_1, \ldots, \varphi_n\}$ be an incongruent set relative to the semantic state space $G$. By definition,

$$\Pr\left[\bigwedge_{i=1}^n \varphi_i\right] = 0, \text{ and } \Pr[\varphi_i] > 0 \text{ for all } i.$$

Equivalently, the truth sets $A_i := \{\omega : \omega \vDash \varphi_i\}$ satisfy

$$\mu\left(\bigcap_{i=1}^{n} A_i\right) = 0.$$

For Boolean indicators $f_i = \mathbf{1}_{A_i}$, this implies

$$\mathbb{E}\left[\prod_{i=1}^{n} f_i\right] = 0.$$

In particular, for any distinct $i, j$,

$$\mathbb{E}[f_i f_j] \leq \min(p_i, p_j),$$

with strict inequality whenever $A_i \cap A_j = \emptyset$. Thus incongruence enforces **negative correlation pressure** among alternatives.

### 5.5.2 Covariance Suppression Lemma

Recall the centered encodings

$$\bar{g}_i = g_i - \mathbb{E}[g_i], \, g_i = 2f_i - 1.$$

*Definition 5.6a (Width / overlap bound)*

Let $g_1, \ldots, g_n : G \to \{0,1\}$ be Boolean functions on the semantic state space $G$. Define the **overlap count** at state $x \in G$ by

$$S(x) := \sum_{i=1}^{n} g_i(x).$$

The family $\Phi = \{g_1, \ldots, g_n\}$ is said to have **width at most $w$** (or $w$-**bounded overlap**) if

$$S(x) \leq w \text{ for all } x \in G.$$

Equivalently, no semantic state satisfies more than $w$ members of the family simultaneously.

**Remarks.**

1. This is a purely combinatorial constraint on the family and is independent of incongruence.
2. If $\Phi$ is incongruent in the sense that $\bigwedge_{i=1}^{n} g_i$ is unsatisfiable, then necessarily $w \leq n-1$, but the converse need not hold. Width is strictly stronger than global incongruence when $w$ is much smaller than $n-1$.

*Definition 5.6b (Uniform width for subfamilies)*

We say $\Phi$ has **uniform width at most $w$** if for every nonempty $J \subseteq \{1, \ldots, n\}$,

$$S_J(x) := \sum_{i \in J} g_i(x) \leq w \text{ for all } x \in G.$$

Equivalently, every subfamily also has overlap bounded by $w$.

*Remark.* Uniform width is equivalent to width for the full family, because $S_J(x) \leq S(x)$ for all $J$. In particular, assuming $S(x) \leq w$, will automatically lead to $S_J(x) \leq w$ for all subfamilies $J$.

*Lemma 5.6 (Quantitative covariance suppression via width)*

Let $J \subseteq \{1, \ldots, n\}$ with $|J| \geq 2$, and let $f_i: G \to \{0,1\}$ be indicator functions with $p_i = \mathbb{E}[f_i]$.

Let
$$S_J(\omega) := \sum_{i \in J} f_i(\omega), \quad w(J) := \operatorname*{ess\,sup}_{\omega} S_J(\omega).$$

Then the following bounds hold:

$$\sum_{i<j \in J} \mathbb{E}[f_i f_j] \leq \frac{w(J)-1}{2} \sum_{i \in J} p_i, \qquad (5.9)$$

and consequently

$$\sum_{i<j \in J} \operatorname{Cov}(f_i, f_j) \leq \frac{w(J)-1}{2} \sum_{i \in J} p_i - \sum_{i<j \in J} p_i p_j. \qquad (5.10)$$

**Proof.**

By definition of width, $S_J(\omega) \leq w(J)$ almost surely. Since $S_J(\omega) \in \{0, 1, \ldots, w(J)\}$, we have the pointwise inequality

$$S_J(\omega)(S_J(\omega) - 1) \leq (w(J) - 1) S_J(\omega) \text{ for all } \omega \in G.$$

Taking expectations yields

$$\mathbb{E}[S_J(S_J - 1)] \leq (w(J) - 1) \mathbb{E}[S_J].$$

Expanding both sides,

$$S_J(S_J - 1) = \sum_{i \neq j \in J} f_i f_j = 2 \sum_{i<j \in J} f_i f_j,$$

and $\mathbb{E}[S_J] = \sum_{i \in J} p_i$. Substituting gives

$$2 \sum_{i<j \in J} \mathbb{E}[f_i f_j] \leq (w(J) - 1) \sum_{i \in J} p_i,$$

which proves (5.9). Subtracting $\sum_{i<j} p_i p_j$ from both sides yields (5.10). ∎

Writing $g_i := f_i - p_i$, note that $\text{Cov}(g_i, g_j) = \text{Cov}(f_i, f_j)$; thus the same bounds apply to the covariance matrix $\Sigma_{ij} = \mathbb{E}[g_i g_j]$.

This shows that incongruence suppresses positive covariance across alternatives.

### 5.5.3 Energy Budget for Incongruent Families

We now combine covariance suppression with the matrix Poincaré inequality.

The inequalities hold for arbitrary families of sentences. The hypothesis of incongruence is not used to strengthen the analytic bound, but to justify its semantic interpretation: only incongruent families represent mutually incompatible alternatives, preventing trivial reuse of variance across jointly realizable sentences.

*Theorem 5.7* (Joint variance suppression for families of sentences)

Let $\Phi = \{\varphi_1, \ldots, \varphi_n\}$ be a finite family of sentences, not necessarily incongruent. For each $i$, let $f_i = \mathbf{1}_{A_i}$ be the indicator of the truth set $A_i = \{\omega : \omega \models \varphi_i\}$, with $p_i = \mathbb{E}_\pi[f_i]$.

Define the aggregate semantic observable

$$S_\Phi(\omega) := \sum_{i=1}^n f_i(\omega),$$

and the width

$$w(\Phi) := \text{ess sup}_{\omega \in \mathcal{M}} S_\Phi(\omega).$$

Then the following statements hold.

**(1) Marginal uncertainty bounds.**

For each $i$,

$$K(\varphi_i) \leq \min\left\{1, \frac{E(\varphi_i)}{2\lambda} \log_2\left(\frac{2e\lambda}{E(\varphi_i)}\right)\right\},$$

and summing over $i$ yields

$$\sum_{i=1}^n K(\varphi_i) \leq \sum_{i=1}^n \min\left\{1, \frac{E(\varphi_i)}{2\lambda} \log_2\left(\frac{2e\lambda}{E(\varphi_i)}\right)\right\}.$$

These bounds are purely componentwise and do not depend on joint satisfiability.

**(2) Counting inequality.**

For any $\kappa \in (0,1]$, there exists a constant $c(\kappa) > 0$ such that

$$|\{i: K(\varphi_i) \geq \kappa\}| \leq \frac{1}{c(\kappa)} \sum_{i=1}^{n} E(\varphi_i).$$

This inequality also holds for arbitrary families and depends only on marginal energies.

**3) Joint variance suppression via width.**

The aggregate variance satisfies

$$\mathrm{Var}_\pi(S_\Phi) \leq w(\Phi) \cdot \sum_{i=1}^{n} p_i - \left(\sum_{i=1}^{n} p_i\right)^2.$$

Equivalently, expanding $(\sum_{i=1}^{n} p_i)^2 = \sum_{i=1}^{n} p_i^2 + \sum_{i \neq j} p_i p_j$, we obtain

$$\mathrm{Var}_\pi(S_\Phi) \leq w(\Phi) \cdot \sum_{i=1}^{n} p_i - \sum_{i=1}^{n} p_i^2 - \sum_{i \neq j} p_i p_j.$$

Equivalently, the total covariance among the family is quantitatively constrained by the width.

In particular, if $\Phi$ is incongruent, then $w(\Phi) \leq n - 1$, and no semantic state can realize all alternatives simultaneously. Consequently, large aggregate variance cannot arise solely from reusing a single highly correlated component across many observables.

**Interpretation.**
The variance bound in part (3) depends only on the width $w(\Phi)$, not on incongruence itself. Incongruence matters indirectly: it enforces $w(\Phi) \leq n - 1$, ruling out the trivial case in which all commitments are simultaneously satisfiable. Other structural constraints may also yield small width without incongruence. Thus, the role of incongruence is to guarantee bounded overlap, not to produce additional variance suppression beyond that implied by width.

**Proof.** For each $i$, let $f_i = \mathbf{1}_{A_i}$ be the indicator of the truth set $A_i = \{\omega: \omega \vDash \varphi_i\}$, and let $g_i = 2f_i - 1 \in \{-1, +1\}$, $\bar{g}_i = g_i - \mathbb{E}[g_i]$, as in Section 5.5.2.

**Step 1: Summed semantic uncertainty bound.**
The single-sentence semantic uncertainty inequality (Theorem 5.4) applies to each $\varphi_i$ individually, yielding

$$K(\varphi_i) \leq \min\left\{1, \frac{E(\varphi_i)}{2} \log_2\left(\frac{2e}{E(\varphi_i)}\right)\right\}.$$

Summing over $i = 1, \ldots, n$ gives the first displayed inequality. This step is componentwise and does not require incongruence.

**Step 2: Counting inequality.**

Fix $\kappa \in (0,1]$ (the case $\kappa > 1$ is trivial since $K(\varphi_i) \leq 1$). Define the function

$$F(E) := \min\left\{1, \frac{E}{2}\log_2\left(\frac{2e}{E}\right)\right\}, \quad E \in [0,2],$$

where we use that $E(\varphi_i) \in [0,2]$ under the normalization used in Section 5.

Consider $h(E) = \frac{E}{2}\log_2(\frac{2e}{E})$ on $(0,2]$. Writing in natural logs,

$$h(E) = \frac{E}{2\ln 2}\ln\left(\frac{2e}{E}\right), \quad h'(E) = \frac{1}{2\ln 2}\left(\ln\left(\frac{2e}{E}\right) - 1\right).$$

Hence $h'(E) \geq 0$ for all $E \in (0,2]$ with equality only at $E = 2$. Thus $h$ is continuous and strictly increasing on $(0,2]$ with $h(E) \downarrow 0$ as $E \downarrow 0$. Therefore, for every $\kappa \in (0,1]$ there exists a unique $c(\kappa) \in (0,2]$ such that

$$h(c(\kappa)) = \kappa.$$

Equivalently, $F(E) \geq \kappa$ if and only if $E \geq c(\kappa)$.

Now let

$$S_\kappa := \{i \in \{1, \ldots, n\} : K(\varphi_i) \geq \kappa\}.$$

If $i \in S_\kappa$, then $K(\varphi_i) \geq \kappa$ and the bound from (1) implies

$$\kappa \leq K\varphi_i \leq F(E(\varphi_i))$$

and hence $E(\varphi_i) \geq c(\kappa)$. Summing over $i \in S_\kappa$ gives

$$\sum_{i \in S_\kappa} E(\varphi_i) \quad |S_\kappa| \ c(\kappa).$$

Since $\sum_{i \in S_\kappa} E(\varphi_i) \leq \sum_{i=1}^n E(\varphi_i)$, we obtain

$$|S_\kappa| \leq \frac{1}{c(\kappa)}\sum_{i=1}^n E(\varphi_i),$$

which is the claimed counting inequality. This step does not require incongruence.

**Step 3 (Width implies aggregate-variance tightening).**

Let $S(\omega) = \sum_{i=1}^n f_i(\omega)$. By definition of width, $S(\omega) \leq w(\Phi)$ almost surely. Since $S(\omega)$ takes nonnegative integer values, we have the pointwise inequality

$$S(\omega)^2 \leq w(\Phi)\, S(\omega) \text{ for all } \omega.$$

We compute the variance via

$$\text{Var}(S) = \mathbb{E}[S^2] - \mathbb{E}[S]^2$$

Since each $f_i$ is an indicator function, we have $f_i(\omega)^2 = f_i(\omega)$ for all $\omega$, and hence

$$S(\omega)^2 = \sum_{i=1}^n f_i(\omega) + 2\sum_{i<j} f_i(\omega)f_j(\omega)$$

Taking expectations yields $\mathbb{E}[S^2] \leq w(\Phi)\mathbb{E}[S]$. Writing $\mathbb{E}[S] = \sum_i p_i$, we obtain

$$\text{Var}(S) = \mathbb{E}[S^2] - \mathbb{E}[S]^2 \leq w(\Phi)\sum_{i=1}^n p_i - \left(\sum_{i=1}^n p_i\right)^2.$$

Finally, since $\sum_i g_i = S - \sum_i p_i$, we also have

$$\text{Var}\left(\sum_{i=1}^n g_i\right) = \text{Var}(S) = \mathbf{1}^\top \Sigma \mathbf{1},$$

**Example 5.7a (same marginals and energies; width changes aggregate variance)**

This example illustrates how aggregate variance depends on overlap and correlation structure through the width parameter, rather than isolating any effect specific to incongruence.

Fix integers $m \geq k$ and choose $n \leq 2^k$. Let $G = \{0,1\}^m$ with uniform measure. For each pattern $a \in \{0,1\}^k$, let

$$A_a := \{\omega \in G : (\omega_1, \ldots, \omega_k) = a\},$$

and define $f_a = \mathbf{1}_{A_a}$. Then each event has the same marginal

$$p := \mathbb{E}[f_a] = 2^{-k},$$

and each predicate has the same boundary sensitivity (hence the same energy $E$) because it depends only on the first $k$ coordinates in the same way (a fixed $k$-bit cylinder set). In particular, all $K(\varphi_a)$ are identical (as functions of $p$) and all $E(\varphi_a)$ are identical by symmetry of coordinates.

Now consider two families of size $n$:

- **Incongruent / low-width family (mutually exclusive):** pick $n$ distinct patterns $a_1, \ldots, a_n$ and let $\Phi_{\text{excl}} = \{\varphi_{a_1}, \ldots, \varphi_{a_n}\}$.
  Then the sets $A_{a_i}$ are disjoint, so $w(\Phi_{\text{excl}}) = 1$ and the family is incongruent (indeed pairwise incompatible). Theorem 5.7(B) gives

  $$V(\Phi_{\text{excl}}) \leq 1 \cdot (np) - (np)^2.$$

- **Non-incongruent / high-width family (duplicates):** let $\Phi_{\text{dup}} = \{\varphi_{a_1}, \ldots, \varphi_{a_1}\}$ consist of the same sentence repeated $n$ times.

Then marginals $p$, entropies $K$, and energies $E$ are identical to the first family *sentence-by-sentence*, but the width is $w(\Phi_{\text{dup}}) = n$, and

$$\sum_{i=1}^{n} f_i = n f_{a_1} \Rightarrow V(\Phi_{\text{dup}}) = n^2 \operatorname{Var}(f_{a_1}) = n^2 \, p(1-p).$$

**Clarification:**
Although $\Phi_{excl}$ is incongruent while $\Phi_{dup}$ is not, the variance gap between $\operatorname{Var}(S_{\text{excl}}) = np(1-p)$ and $\operatorname{Var}(S_{\text{dup}}) = n^2 p(1-p)$ arises entirely from differences in correlation structure: $\Phi_{excl}$ consists of pairwise disjoint (hence uncorrelated) indicators, while $\Phi_{dup}$ consists of perfectly correlated duplicates. The example does not isolate incongruence as an independent source of variance suppression; rather, it demonstrates how bounded overlap prevents variance from being reused along a single shared direction.

**Conclusion:** the two families have identical marginals $p_i$, identical entropies $K(\varphi_i)$, and identical energies $E(\varphi_i)$ for each $i$, yet their **aggregate variance** differs by a factor $\Theta(n)$–$\Theta(n^2)$ because width controls whether variance can be "reused" along a shared direction. This is precisely the sense in which incompatibility/overlap structure yields a genuine tightening beyond marginal bounds. The contrast here illustrates the effect of overlap structure rather than incongruence per se: the incongruent family has bounded width, while the duplicated family has maximal width despite identical marginals.

### 5.5.4 Interpretation

This result formalizes the intuition that **incongruent alternatives compete for semantic energy**. One cannot maintain many highly informative but mutually incompatible alternatives at arbitrarily low energy.

Thus incongruence does not merely prevent joint satisfiability; it enforces a **quantitative tradeoff**:

- More informative alternatives $\Rightarrow$ higher total semantic energy.
- Low total energy $\Rightarrow$ collapse of knowledge across alternatives.

### 5.5.5 Conceptual Consequence

This tightening explains why expressive systems exhibit either:

- few sharp alternatives, or
- many vague ones,
  but never many sharp incompatible alternatives without increased semantic complexity.

Paradox appears precisely when this energy budget is violated by forcing incompatible alternatives into a single determinate description.

### 5.5.6 Incongruence Tightening Principle.
In any incongruent family of sentences, semantic energy is a conserved resource. Maintaining many informative incompatible alternatives requires proportionally higher semantic energy; enforcing low energy forces informational collapse.

## 6 DISCUSSION

This section discusses the conceptual and technical implications of the results, clarifies their scope, and positions semantic uncertainty as a foundational constraint on semantic and knowledge-representation systems.

## 6.1 What Has Been Shown

The results of Sections 3–5 establish three progressively stronger claims.

First, self-referential semantic paradoxes can be transformed into **incongruent normal form**, eliminating explicit self-reference while preserving semantic obstruction. Paradox is revealed as structured incompatibility rather than semantic failure.

Second, **semantic completeness is incompatible with knowledge**. Systems in which all sentences are globally determined admit no informative statements. Incompleteness is therefore not merely unavoidable but necessary.

Third, this necessity admits a **quantitative formulation**. Semantic energy bounds semantic knowledge content, yielding a genuine uncertainty principle: semantic determinacy, informativeness, and spectral simplicity cannot be simultaneously optimized. For families of sentences, incongruence tightens this bound via a matrix inequality that enforces redistribution of semantic energy across incompatible alternatives.

Taken together, these results move incompleteness from a negative limitation to a positive structural requirement.

## 6.2 Relation to Gödel and Heisenberg (Clarified)

Gödel's incompleteness theorem showed that no sufficiently expressive axiomatic system can be both complete and consistent. Heisenberg's uncertainty principle showed that certain physical observables cannot be simultaneously determined with arbitrary precision.

The present work establishes an analogous boundary, but at the level of **semantic representation** rather than proof or measurement.

- Gödel constrains *formal provability*.
- Heisenberg constrains *physical observability*.
- Semantic uncertainty constrains *meaning and knowledge*.

While not as fundamental as Gödel's result about provability or Heisenberg's about measurement, the semantic uncertainty bound provides a quantitative constraint on semantic representation that complements these classical impossibility results. Importantly, this is not an analogy but a structural similarity: all three results identify a tradeoff that cannot be engineered away. The semantic uncertainty bound shows that attempts to eliminate incompleteness do not merely fail; they collapse knowledge quantitatively.

## 6.3 Why Incongruence Is Essential (Not Optional)

A key contribution of this work is identifying **incongruence** as the minimal structure that realizes semantic uncertainty.

Incongruent sets:

- preserve classical semantics locally,
- forbid global semantic collapse,
- force redistribution of semantic energy across alternatives.

Without incongruence, semantic energy can concentrate into a single determinate assignment, and knowledge vanishes. Thus incongruence is not a defect or a symptom of poor modeling; it is the mechanism by which semantic degrees of freedom are preserved.

This explains why paradox persists in expressive systems and why attempts to fully eliminate it repeatedly reintroduce expressive limitations elsewhere.

### 6.4 Implications for Knowledge Representation and Reasoning

The results have direct implications for the design of logical and computational systems.

1. **Consistency enforcement has a cost.**
   Enforcing global semantic consistency reduces semantic energy and collapses knowledge content.
2. **Incompleteness is a design constraint.**
   Knowledge-representation systems must preserve unresolved alternatives in a principled way.
3. **Inconsistency tolerance is structurally justified.**
   Systems that tolerate incompatible but locally coherent assertions are not weakened; they are knowledge-capable by necessity.

The semantic uncertainty bound provides a quantitative way to reason about these tradeoffs.

### 6.5 Matrix Uncertainty and Multi-Sentence Knowledge

The matrix formulation developed in Section 5 shows that semantic uncertainty is not merely a per-sentence phenomenon. For families of sentences, total variance—and hence total knowledge—is bounded by total semantic energy.

In incongruent families, covariance suppression prevents variance from being shared, forcing a tradeoff between the number of informative alternatives and the energy budget available. This provides a formal explanation for why systems cannot maintain many sharp, mutually incompatible hypotheses without increased semantic complexity.

This result aligns with empirical behavior in knowledge systems, where either few hypotheses are held strongly or many are held weakly, but not both simultaneously.

### 6.6 Scope and Limitations

The quantitative bounds depend on modeling semantic states as a finite abelian group, a standard assumption in theoretical computer science. This provides access to Fourier-analytic tools and Dirichlet forms.

The qualitative results—semantic incompleteness as necessary for knowledge and paradox as a boundary phenomenon—do not depend on this assumption. The group structure is required only for the quantitative uncertainty bounds.

Future work may explore alternative semantic geometries and corresponding uncertainty relations.

### 6.7 Foundational Value

The foundational value of this work lies in establishing a **structural constraint on semantic representation**:

**Semantic systems cannot be made fully determinate without quantitatively destroying knowledge.**

This result shifts the interpretation of paradox, incompleteness, and inconsistency. They are not failures to be repaired but signals that a system has reached the boundary of semantic compression.

Much as Gödel's theorem reshaped expectations about axiomatization and Heisenberg's principle reshaped expectations about measurement, semantic uncertainty provides a quantitative constraint on semantic representation.

### 6.8 Implications for knowledge representation and statistical semantic systems.

Although our results are purely semantic and do not model neural architectures, they suggest a general representational tension relevant to statistical semantic systems, including modern ML language models. The semantic uncertainty bound formalizes a tradeoff between determinacy and informativeness in finite semantic-state models: pushing representations toward global determinacy collapses entropy and thereby eliminates semantic differentiation. This provides a theoretical lens—independent of any particular learning algorithm—for why systems that represent meaning may need to tolerate controlled incompleteness to remain informative. Separately, incongruent normal form can be viewed as a structural representation of conflict: rather than treating inconsistency as a failure state, it isolates the precise locus of joint unsatisfiability among locally coherent constraints, a perspective aligned with inconsistency-tolerant knowledge representation and belief-maintenance settings. We view these connections as suggestive; establishing operational consequences for trained models is an interesting direction for future work.

### 6.9 Outlook

The framework introduced here opens several directions for future work, including refined energy measures, alternative semantic state spaces, and applications to inconsistency-tolerant reasoning and knowledge-based systems. More broadly, it suggests that semantic incompleteness should be treated as a first-class design principle rather than a technical inconvenience.

## 7 CONCLUSION

We have shown that self-referential semantic paradoxes can be transformed into incongruent normal form, isolating incompatibility as a structural property rather than a failure of truth assignment. Using this framework, we proved that semantic completeness eliminates informativeness and therefore cannot support knowledge.

We further established a quantitative semantic uncertainty principle, bounding the knowledge content of a sentence by its semantic energy. This bound extends to families of sentences via a matrix inequality and tightens under incongruence, yielding a conserved energy budget across incompatible alternatives.

These results identify semantic incompleteness as a necessary design constraint rather than a technical limitation. Attempts to enforce total semantic determinacy quantitatively collapse knowledge.

Future work includes refining semantic energy measures, extending uncertainty bounds to other semantic geometries, and applying the framework to inconsistency-tolerant reasoning systems.

## 8   RELATED WORK

**Paraconsistency and dialetheism.** A long tradition treats semantic paradox by tolerating inconsistency, often via paraconsistent logics in which explosion fails. Priest's dialetheic view and the Logic of Paradox (LP) provide a canonical framework in which some contradictions can be true without trivializing the theory [14], [15]. Our approach differs in that we do not change consequence relations or endorse true contradictions; instead, we transform self-reference into a finite family of classically interpretable constraints whose joint incompatibility is made explicit as incongruence.

**Impossible worlds and non-classical model structures.** Another family of approaches models inconsistent or incomplete information using enriched semantic spaces, including impossible worlds semantics and related non-normal modal semantics [16] - [18]. These frameworks also support "local coherence without global satisfaction," but typically by expanding the space of admissible worlds. In contrast, incongruent normal form keeps classical local satisfaction and isolates incompatibility at the level of jointly satisfiable constraints, serving as a normal-form representation rather than an alternative semantics.

**Revision theory and circular definitions.** Gupta and Belnap's revision theory analyzes circular definitions and semantic predicates via iterated revision sequences rather than fixed truth assignments [19]. From that viewpoint, paradox manifests as non-convergence or oscillation under revision. Our INF transformation can be viewed as extracting the finite constraint cycle responsible for such non-stabilization and expressing it as an explicit incompatibility set; the quantitative bounds in Section 5 then provide an orthogonal perspective by measuring how semantic differentiation is limited by sensitivity (energy) within finite semantic-state models. Dunn's four-valued and First-Degree Entailment (FDE) frameworks similarly model information states with local consistency and global incompleteness, but do so by enriching truth values rather than by isolating incompatibility as a structural normal form.